\documentclass[12pt, manuscript,screen]{acmart}

\usepackage[utf8]{inputenc}
\usepackage{arabtex}
\usepackage{utf8}
\setcode{utf8}
\novocalize
\makeatletter
\let\prism@latex@error\@latex@error
\def\@latex@error#1#2{\PackageWarning{LaTeX}{#1}}
\AddToHook{begindocument/end}{\let\@latex@error\prism@latex@error}
\makeatother
\usepackage{tikz}
\usetikzlibrary{positioning, arrows.meta}
\usepackage{tcolorbox}
\tcbuselibrary{breakable}
\usepackage{listings} 
\usepackage{pgfplots}
\usepackage{pgf-pie}
\usepackage{multirow}

\usepackage{xcolor}

\AtBeginDocument{%
  }

\acmDOI{XXXXXXX.XXXXXXX}
\acmConference[Journal of Computer and Communications]{Journal of Computer and Communications}{2025}{Online}
\acmISBN{978-1-4503-XXXX-X/2018/06}

\begin{document}

\title{BOUTEF: A Multilingual Corpus for Fake News in North Africa — Language as a Weapon}
\author{Kamel Smaïli}
\orcid{0000-0002-4237-7303}
\email{smaili@loria.fr}
\affiliation{%
  \institution{Loria, CNRS, Université de Lorraine}
  \city{Vandoeuvre-Lès-Nancy}
  \country{France}
}

\author{Yassine Toughrai}
\orcid{0009-0005-6188-9100}
\email{yassine.toughrai@loria.fr}
\affiliation{%
  \institution{Loria, CNRS, Université de Lorraine}
  \city{Vandoeuvre-Lès-Nancy}
  \country{France}
}

\author{Amina Laggoun}
\orcid{0009-0006-7698-1687}
\email{amina.laggoun@loria.fr}
\affiliation{%
  \institution{Loria, CNRS, Université de Lorraine}
  \city{Vandoeuvre-Lès-Nancy}
  \country{France}
}
\affiliation{%
  \institution{Université Internationale de Rabat}
  \city{Rabat}
  \country{Morocco}
}

\author{David Langlois}
\orcid{0000-0001-7277-3668}
\email{david.langlois@loria.fr}
\affiliation{%
  \institution{Loria, CNRS, Université de Lorraine}
  \city{Vandoeuvre-Lès-Nancy}
  \country{France}
}

\renewcommand{\shortauthors}{Smaïli.}

\begin{abstract}
The rapid spread of fake news on social media has become a major challenge, particularly in multilingual and under-resourced contexts such as North Africa. In this paper, we introduce BOUTEF, a large-scale multilingual corpus designed to study the propagation, characteristics, and impact of fake news in Algeria and Tunisia. The corpus integrates three complementary components: fake narratives, genuine narratives, and associated user-generated comments, along with verified debunking information. It covers a wide range of languages and linguistic varieties, including Modern Standard Arabic, Algerian and Tunisian dialects, Arabizi, French, English, and code-switchd languages.
Building on this resource, we conduct a comprehensive empirical analysis combining quantitative and qualitative approaches. We examine thematic distributions, linguistic and rhetorical strategies, sentiment patterns, and social engagement dynamics. Statistical analyses reveal significant associations between thematic categories and message veracity, as well as strong correlations between user engagement and the visibility of fake content. Our findings show that fake news tends to rely on emotionally charged narratives, sensational framing, and hybrid linguistic practices that enhance virality and audience engagement. In contrast, debunking content adopts a more factual and verification-oriented style.
Furthermore, a comparative analysis between Algeria and Tunisia highlights both shared dynamics and country-specific characteristics shaped by sociopolitical contexts. The results emphasize the role of multilingualism and informal language practices in the diffusion and reception of misinformation. By providing a rich, annotated, and publicly available dataset, this work contributes to advancing research on fake news detection, low-resource language processing, and the understanding of information disorders in complex linguistic environments.
\end{abstract}

\keywords{\it \ \ Fake news, Under-resourced languages, misinformation, Arabic dialects Multi-lingual corpus, code-switching}

\settopmatter{printacmref=false}
\maketitle

\section{Introduction}
Attention war and cognitive war provide an important conceptual framework for understanding how disinformation captures public attention and shapes perceptions, beliefs, and decision-making. In the digital age, fake news has become one of the most visible and disruptive forms of disinformation. Social media platforms play a central role in its diffusion, as their engagement-driven architectures can amplify sensational, emotional, or polarizing content and reinforce echo chambers \cite{Ferraz2024}. Beyond attracting attention, such content can shape beliefs, distort perceptions of reality, and influence public opinion. In this sense, fake news lies at the intersection of attention dynamics and cognitive influence, making it a major challenge for contemporary information ecosystems.

Although the global impact of fake news has been widely documented, its dynamics remain underexplored in North African contexts. This gap is especially important given the region's linguistic diversity, complex political transitions, and intensive use of social media. North African online discourse is inherently multilingual and frequently combines dialectal Arabic, Modern Standard Arabic, Arabizi, French, English, and code-switching practices. These characteristics make the region particularly relevant for studying how misinformation is produced, circulated, and received across heterogeneous linguistic and cultural settings.

An initial version of the BOUTEF corpus was introduced in \cite{smaili:hal-04578297}, but it remained relatively small. In this article, we present the definitive and substantially expanded version of BOUTEF, a multilingual corpus that includes fake and genuine narratives, user comments, and debunking information collected from multiple social media platforms. By capturing the linguistic and communicative diversity of Algeria and Tunisia, the corpus provides a valuable resource for analyzing misinformation in under-resourced and hybrid language environments.

Building on this resource, we conduct a comprehensive study of fake news in Algeria and Tunisia. Our analysis examines thematic distributions, rhetorical and linguistic strategies, user engagement patterns, and sociopolitical specificities. Particular attention is paid to hybrid forms such as Arabizi, which play an important role in online interaction and may contribute to the spread and reception of misinformation. We also compare the two national contexts in order to identify both shared regional dynamics and country-specific characteristics.

This article makes three main contributions. First, it presents the definitive version of BOUTEF, a large-scale corpus for the study of multilingual fake news in North Africa. Second, it provides an extensive empirical analysis of misinformation across thematic, linguistic, and engagement-related dimensions. Third, it offers a comparative perspective on Algeria and Tunisia, highlighting how local historical, cultural, and political contexts shape misinformation ecosystems.

The remainder of the article is organized as follows. Section 2 reviews related work. Sections 3 and 4 introduce the BOUTEF corpus and the methodology adopted in this study. Section 5 examines methods of manipulation and fake news fabrication. Section 6 presents the main quantitative characteristics of the corpus. Section 7 investigates correlations between fake news, social media metrics, and algorithmic amplification. Section 8 compares the geographical origins of fake news targeting Algeria versus Tunisia. Section 9 measures linguistic diversity across the different message types in the BOUTEF corpus. Finally, Section 10 discusses the implications of our findings and outlines directions for future research.

\section{Related work}
\begin{figure}[htbp]
    \centering
    \includegraphics[width=0.45\textwidth]{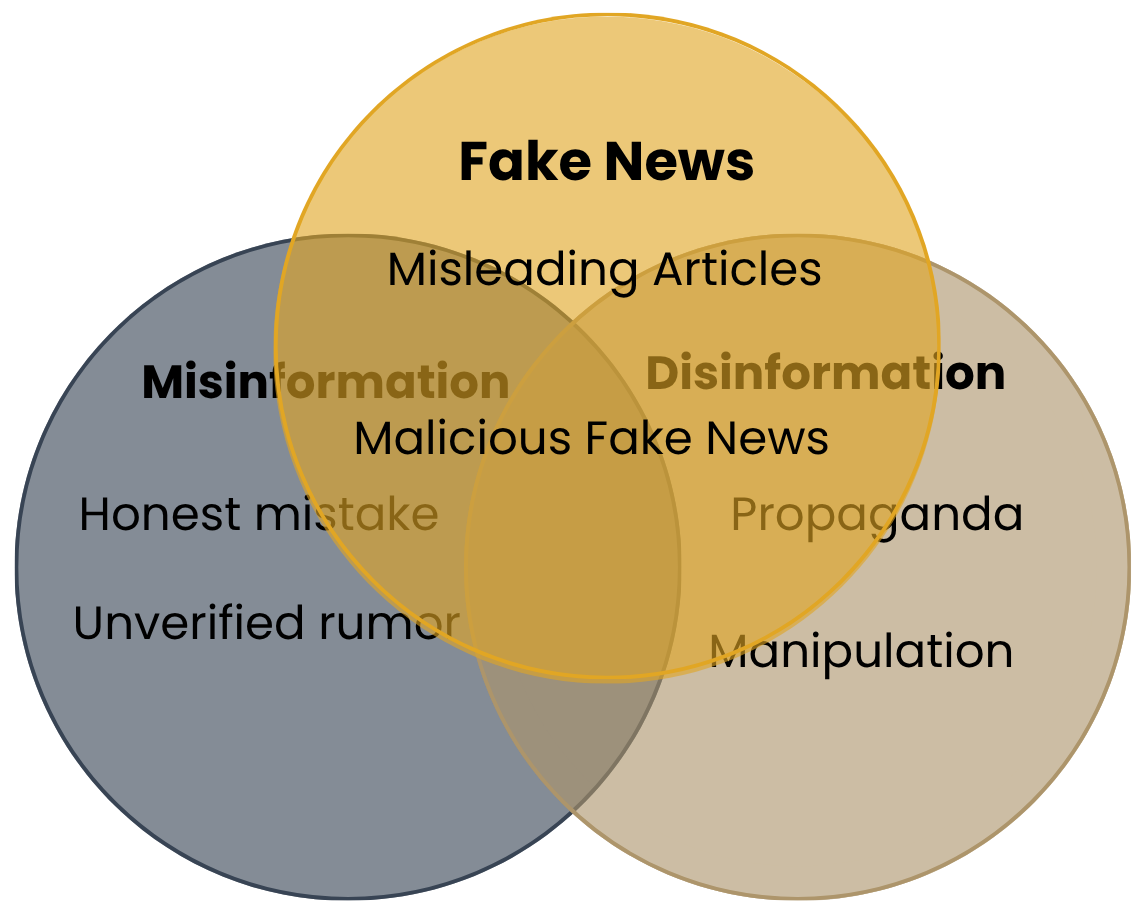}
    \caption{Distinction between Fake news, misinformation and disinformation.}
\label{Mis}
\end{figure}

First, we define the term of fake news and its related concepts in connection with attention war. Within this perspective, the distinction between misinformation and disinformation~\cite{lecheler2022disinformation} is crucial for understanding how misleading content captures attention and shapes public perceptions. Misinformation refers to unintentional inaccuracies, while disinformation denotes deliberate falsehoods, often strategically crafted to influence public opinion. Fake news, a broader and popular term, encompasses fabricated or exaggerated stories that may fall into either category depending on intent. In the context of attention war, such content is designed not only to mislead but also to maximize visibility, emotional reaction, and circulation. Figure \ref{Mis} illustrates these overlaps.

Analysis of the BOUTEF corpus shows that while both forms exist, politically motivated content is most often disinformation, reflecting deliberate manipulation of public opinion. Its persistence, despite counter-evidence, shows how difficult disinformation is to combat and the importance of fact-checking and media literacy.
\label{related_works}
Before reviewing existing datasets and studies, it is important to highlight the broader context in which fake news propagates. Social media platforms have increasingly become the main channels for news consumption, influencing public opinion at an unprecedented scale. This section surveys prior research on fake news detection, focusing on datasets, languages, and labeling strategies. By situating our work within this landscape, we clarify the gaps that our BOUTEF corpus aims to fill. As summarized in Table~\ref{tab:fake_news_corpora}, large-scale English datasets such as FakeNewsNet, FEVER, and LIAR are available, whereas Arabic resources (e.g., DART, Saadany Satirical) remain limited. Although the table does not list all existing corpora, none incorporate user comments or debunking information, two elements that constitute the core novelty of our BOUTEF corpus.
\begin{table}[h]
\centering
\scriptsize

\begin{tabular}{|p{3cm}|p{2cm}|p{2cm}|p{2cm}|p{2cm}|}
\hline
\textbf{Name / Ref.} & \textbf{Lang.} & \textbf{Content} & \textbf{Labels} & \textbf{Size / Year} \\
\hline
FakeNewsNet \cite{shu2020} & Eng & Text, images, social & 2-way & 23.9k/ 2020 \\
FEVER \cite{Thorne2018} & Eng & Text claims & 3-way & 185k/ 2018 \\
LIAR \cite{wang2017liar} & Eng & Short statements & 6-way & 12.8k/ 2017 \\
PHEME \cite{zubiaga2016} & Eng/German & Tweets conv & 3-way & 4.8k/ 2016 \\
Fakeddit \cite{nakamura2020} & Eng & Posts & 2/3/6-way & 1M/ 2020 \\
CREDBANK \cite{Mitra2021} & Eng & Tweets & 5-way & 60M/ 2021 \\
BuzzFace \cite{Williams2018} & Eng & Facebook posts & 4-way & 2.3k/ 2018 \\
COVID-19 FN \cite{cheng2021} & Eng & News articles & 3-way & 6.8k / 2021 \\
MM-COVID \cite{li2020} & Multiling (6) & Articles & 2-way & 11k / 2020 \\
MuMiN \cite{Nielsen2022} & Multiling (41) & Tweets & Fact-checked & 21M / 2022 \\
BanglaFakeNews \cite{hossain2020} & Bangla & Articles & 2-way & 50k / 2020 \\
DART \cite{Rasha2023} & Arabic (5 dia) & Tweets & N/A & 25k / 2023 \\
Satirical \cite{saadany2020fake} & Arabic & Articles & Satire & 3.2k / 2020 \\
YouTube Rumors \cite{Maysoon2019,alkhair2023} & Arabic & Comments & Rumors & 3k / 2019 \\
AFND \cite{khalil2022afnd} & Arabic & News & 3-way & 606k / 2021 \\
AI-Generated \cite{himdi2025arabic} & Arabic & Articles & 2-way & 1.5k / 2022-23 \\
\hline
\end{tabular}
\caption{Representative Fake News Corpora (sorted by language and scope). 
Size is measured in terms of the dataset's content unit. The DART corpus 
covers five Arabic dialect groups: Egyptian, Maghrebi, Levantine, Gulf, 
and Iraqi. Note that the Maghrebi category groups together dialects from 
different countries (e.g., Algerian, Tunisian), unlike BOUTEF which treats them as distinct dialects.}
\label{tab:fake_news_corpora}
\end{table}

\subsection{Social media and fake news}
Social media has become a primary news source for much of the global population, surpassing television \cite{aimeur2023fake}. Its accessibility, speed, and interactivity make it especially popular among younger generations. However, these same features facilitate the rapid spread of false information. Detecting fake news remains challenging, partly due to the lack of comprehensive datasets that jointly capture content, social context, and misinformation categories \cite{sunstein2009,Penny2021,shu2020,aimeur2023fake}. Constructing such datasets requires careful consideration of topic coverage, scale, language, and metadata.

\subsection{English and multilingual datasets}
Existing corpora vary widely in scope. Some target specific events, such as the 2016 US elections \cite{pathak2019breaking} or the COVID-19 pandemic \cite{mahlous2021fake,du2021cross}, while others cover broader domains such as politics \cite{wang2017liar}. Large multi-domain resources span politics, economy, healthcare, sports, and entertainment \cite{khan2019benchmark,khalil2021detecting}. Despite these resources, models often struggle to generalize across domains \cite{khan2019benchmark}.  

Most datasets are in English \cite{pathak2019breaking,khan2019benchmark,wang2017liar}, but others exist in Spanish \cite{posadas2019detection}, Portuguese \cite{monteiro2018contributions,charles2022fakepedia}, German \cite{vogel2019fake}, Chinese \cite{nakamura2020}, Bangla \cite{hossain2020}, and Urdu \cite{amjad2020bend}. Formats and sizes differ: some contain long articles \cite{khan2019benchmark,saadany2020fake}, while others focus on short texts such as claims or tweets \cite{wang2017liar,mahlous2021fake}. Labeling strategies also vary, ranging from binary \cite{mahlous2021fake} to multi-level veracity scales \cite{wang2017liar} or categories such as {\it satire}, {\it misleading}, and {\it true} \cite{khalil2021detecting}.

\subsection{Arabic datasets}

For Arabic, research efforts remain limited. A well-known resource is AFND~\cite{khalil2022afnd} (Arabic FakeNews Dataset), which contains 606,912 publicly available news articles collected from 19 Arab countries and manually verified. AFND is written in Modern Standard Arabic. The DART dataset \cite{Rasha2023} provides 25,000 tweets in five dialects through crowdsourced annotation. In \cite{Shatha2023}, fake news tweets were collected from fact checking platforms such as Anti-Rumors Authority and Misbar. Additional resources include a collection of 3,185 satirical articles \cite{saadany2020fake}, as well as datasets built from YouTube comments related to death rumors about public figures such as Bouteflika, Adel Imam, and Fifi Abdo \cite{Maysoon2019,alkhair2023}. These initiatives highlight the scarcity of resources for under-resourced Arabic dialects compared to English.  

More recent work has introduced datasets designed to study fake news generated by large language models. For example, \cite{himdi2025arabic} includes 500 real articles, 500 fake articles written by humans, and 500 fake articles generated by AI. Similarly, \cite{mohamed2026enhancing} explores data augmentation through automatic methods such as synonym replacement and LLM-based generation. Starting from an initial COVID-19 tweet dataset, the authors double the size of the data and report improved classifier performance, although the resulting dataset does not appear to be publicly available. This growing interest in LLM-generated misinformation corpora reflects a broader shift in the field, as models such as GPT-4 \cite{openai2023gpt4} and LLaMA \cite{touvron2023llama} have demonstrated a strong ability to produce fluent and coherent text at scale. Several studies have begun exploring the linguistic properties of AI-generated fake news and how they differ from human-written content \cite{uchendu2021turingbench}, motivating the construction of datasets that explicitly account for this new category of disinformation.

\subsection{Rethinking fake news datasets: design, limitations, and contributions of BOUTEF}
A major limitation of many existing corpora lies in their conceptual ambiguity: they often fail to distinguish clearly between genuinely fake content and texts that merely discuss, report on, or debunk fake news. This lack of precision reduces their usefulness for fine-grained analysis and can blur the boundary between misinformation and its commentary. In addition, many available resources provide only the fake item itself, without the surrounding interactional context that is crucial for understanding how misinformation is received, contested, and amplified online.

BOUTEF is designed to address these limitations. It explicitly separates three complementary components: {\it fake news articles, user comments, and debunking information}. User comments make it possible to analyze public reactions, perceived credibility, and engagement dynamics, while debunking content provides verified counter-evidence and supports the study of correction mechanisms. We further enrich the resource by including genuine narratives together with their comments, collected from verified websites, which enables direct comparison between false and authentic information.

Building on this design, BOUTEF offers substantial added value as a multilingual resource covering Algerian and Tunisian dialects, French, English, Modern Standard Arabic, and code-switching practices. By combining linguistic diversity, interactional context, and verified debunking, the corpus better reflects the real circulation of misinformation in a part of North Africa. It therefore supports not only fake news detection, but also broader investigations into audience reception, veracity perception, debunking strategies, and the sociolinguistic dynamics of misinformation in under-resourced settings.
\section{Corpus overview}
BOUTEF is composed of two parts: the first concerns fake narratives (BOUTEF-Fake), and the second concerns genuine narratives (BOUTEF-Real). 
\subsection{BOUTEF-Fake}
\begin{figure}[htbp]
    \centering
    \includegraphics[width=0.7\textwidth]{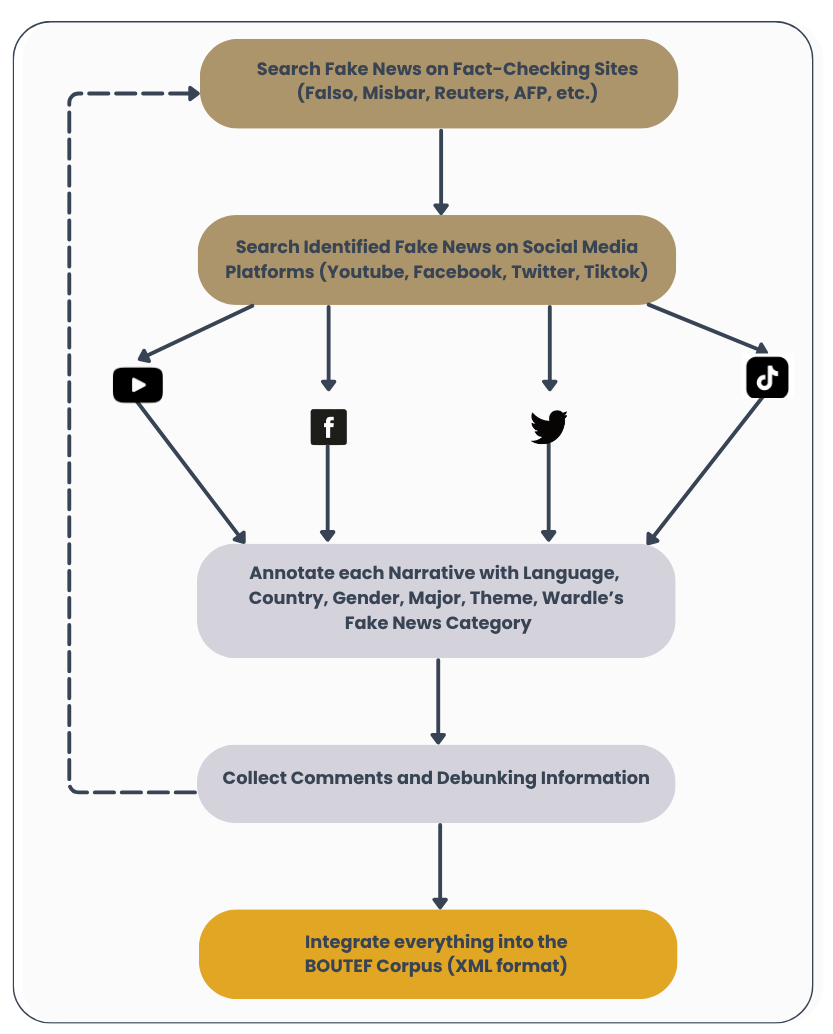}
    \caption{Workflow for constructing the BOUTEF-Fake (iterative feedback with right-angled open rectangle at left).}
    \label{Const}
\end{figure}

The construction of the BOUTEF-fake part (Figure \ref{Const}) follows a step-by-step workflow that combines fact-checking resources, social media platforms, and iterative annotation. The process begins by identifying false information through established fact-checking sites such as Misbar\footnote{\url{https://www.misbar.com/}}, Falso \footnote{\url{https://www.facebook.com/falso.tn/}}, DZ Fake news \footnote{\url{https://www.facebook.com/FakenewsDZ/}}, and AFP. These verified fake news items are then traced on major social platforms including YouTube, Facebook, Twitter/X, and TikTok, where the same narratives are collected to enrich the corpus. Each narrative is subsequently annotated with key metadata such as language, country, gender, major theme, and Wardle’s fake news category \cite{Warde}, ensuring a systematic description of content. In parallel, user comments and debunking information related to these narratives are gathered to capture the surrounding discourse. Finally, all elements are integrated into the BOUTEF corpus in XML format as presented below.\\

\vspace{0.5cm}
\begin{minipage}{\dimexpr\linewidth-4mm\relax}
\noindent
\small
\begin{tcolorbox}[
    colback=white,
    colframe=black,
    boxrule=0.5pt,
    arc=4pt,
    breakable,
    width=\linewidth,
    left=0pt,
    right=0pt,
    boxsep=0pt
]
\lstset{
    language=XML,
    basicstyle=\ttfamily\footnotesize,
    breaklines=true,
    breakatwhitespace=true,
    columns=fullflexible,
    showstringspaces=false,
    escapeinside={\%*}{*\%}
}
\begin{lstlisting}
<Entry>
<CONCERN>ALGERIE</CONCERN>
<TYPE>CHECKED</TYPE>
    <MAJOR_THEME>POLITIQUE</MAJOR_THEME>
    <MINOR_THEME>ECONOMIE</MINOR_THEME>
    <SUBJECT>PAUVRETE ALGERIE</SUBJECT>
    <IMAGE_REFERENCE>Pauvre.png</IMAGE_REFERENCE>
    <MESSAGE>\%*\RL{الجزائر دولة بترولية لكن شعبها يعيش تحت رحمة الفقرهذه امنيات الجزائريين}*\%</MESSAGE>
    <TYPE_MESSAGE>Fake</TYPE_MESSAGE>
    <FAKE_CATEGORY>DIVERTED INFORMATION</FAKE_CATEGORY>
    <LANGUAGE>MSA</LANGUAGE>
    <PUBLICATION_DATE>2024-05-02 00:00:00</PUBLICATION_DATE>
    <COMMENTS_NUMBER></COMMENTS_NUMBER>
    <SITE>Twitter</SITE>
    <NUMBER_FOLLOWERS>4774.0</NUMBER_FOLLOWERS>
    <NUMBER_OF_FORWARD></NUMBER_OF_FORWARD>
    <COUNTRY></COUNTRY>
    <GENDER>S</GENDER>
    <USER></USER>
  </Entry>
\end{lstlisting}
\end{tcolorbox}
\end{minipage}
 \\

A narrative in the BOUTEF corpus is represented as a structured XML-like entry that combines thematic annotation, textual content, and contextual metadata. As illustrated by the example above, each entry typically specifies the country or concern, the source type of the item, its major and minor themes \cite{smaili:hal-04578297}, a short subject label, and the main message itself. It also includes veracity-related information such as the message label (e.g., fake or genuine) and, when relevant, the fake-news category. In addition, the structure records linguistic and platform-level metadata, including the language, publication date, site, engagement indicators, and user-related attributes. This organization makes each narrative not only a text to be analyzed linguistically, but also a documented social-media object that can be studied in relation to topic, source, audience reaction, and circulation context.

The dashed U-shaped arrow in the workflow of Figure \ref{Const} indicates an iterative feedback loop, where the analysis of comments informs and enriches the initial search and annotation phases. This workflow illustrates the data collection and iterative annotation process that creates BOUTEF-fake. In the following, we detail the three main components of this part of the corpus: fake news articles, fake-related comments, and debunking information.

BOUTEF-Fake corpus \cite{smaili:hal-04578297} is composed of three primary components. The first one concerns the  {\bf Fake News Articles} where each article is classified under one of seven “Fake Categories” inspired from the classification of Wardel \cite{Warde}:
\begin{itemize}
\item Diverted Information: Refers to the use of accurate information in a misleading context.
 \item Misleading Content: Involves the use of deceptive or ambiguous language to manipulate perception.
\item Fabricated Content: Encompasses entirely invented narratives or data.
\item Exaggeration or Extension: Distorts or amplifies existing facts to create a false impression.
\item Manipulated Content: Includes alterations or misrepresentations of genuine media, such as images or videos.
\item Satire or Parody: Humorous or ironic content mistaken for factual reporting.
\item Deny the Existence: Outright denial of verifiable events or facts.
\end{itemize}

The second one concerns the {\bf Fake-Related Comments}. These user-generated responses display a range of linguistic styles and languages including Modern Standard Arabic (MSA), Algerian dialect, Tunisian dialect, Arabizi, Tunisian Arabizi, code-switching among Arabic, French, and English, as well as Modern Standard Arabic in Latin script (MARABIZI). In fact the user comments play a crucial role, as individuals share their opinions on various types of news by posting comments, reactions, and expressing skeptical views. In the context of attention war, these comments are not merely reactions: they become signals that amplify, redirect, or contest visibility around specific narratives. They therefore form a key interface between attention capture and cognitive influence, where engagement dynamics can reinforce the circulation of disinformation. As a result, these comments may contain valuable semantic information that can help distinguish between real and fake news \cite{albahar2021hybrid}. To emphasize the significance of fake comments, Zubiaga et al. \cite{zubiaga2018detection} suggested that their method enables the differentiation of user comments. The third component concerns the {\bf Debunking Information (NoFake)} composed by documents and evidence aimed at countering the misinformation spread by the fake articles. These information are extracted from fact-checking website such as: MISBAR, FALSO, DZ Fake news, AFP, REUTERS, etc.
Additional metadata such as geographical tags, publication dates, and creator profiles (where available) enhance the analytical potential of the corpus.
\subsection{BOUTEF-Real}
This second part of BOUTEF focuses on genuine narratives collected in the final stage of the pipeline, using keywords derived from fake posts but targeting verified information. Concretely, we identify the fake claim, extract its central entities and events as search keywords, retrieve reliable reports on the same topic, verify the content, label it as genuine, and integrate it into BOUTEF. For example, if a BOUTEF-Fake post claims that President Tebboune died, we use keywords such as Tebboune visit to Portugal on a later date to retrieve evidence that confirms the information is genuine rather than fake.

As in BOUTEF-Fake, we integrate  the comments associated with the genuine information. These were collected simultaneously with the genuine content, for example from platforms such as Youtube, at the time the comments were published.

In Table \ref{TabStatBou}, we present the count of posts within this corpus containing over 1,09M words. This corpus is associated with 194 images depicting fake content. Images and videos play a crucial role in spreading fake news because they create a powerful first impression that quickly registers in people's minds and often bypasses critical thinking.

\begin{table}[h]
\begin{center}
  \begin{tabular}{cc}
        \hline
        {Number of posts}&{\bf 57095}\\
        \hline
        {Number of words}& {\bf 1,09M}\\
       
        {Number of unique words}&{\bf 104k}\\
        
        {Number of images}&{\bf 194}\\
        \hline
        \end{tabular}      
        \caption{Some figures on the BOUTEF corpus.}
        \label{TabStatBou}
        \end{center}
\end{table}
\subsection{Corpus structure overview}
\begin{figure}[htbp]
    \centering
    \includegraphics[width=\textwidth]{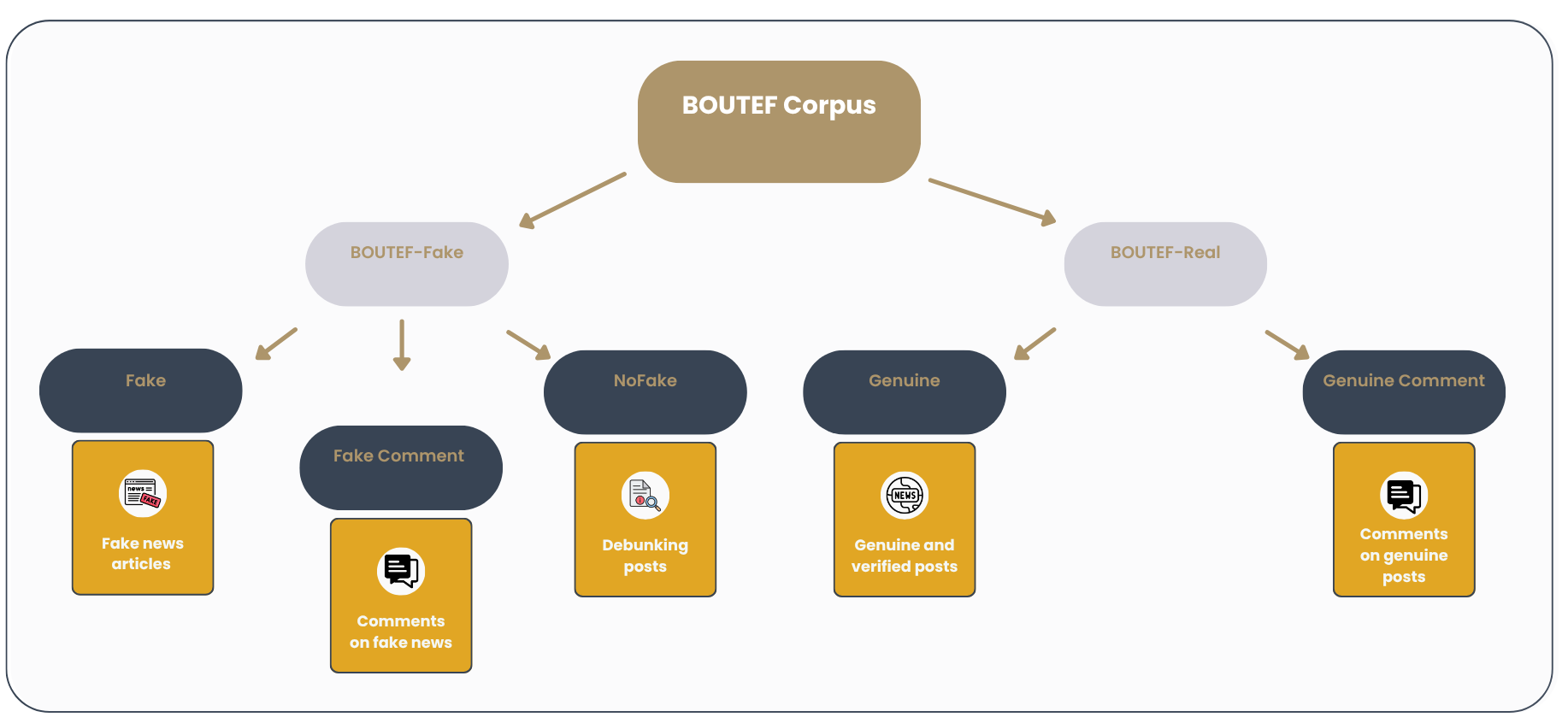}
    \caption{Overall structure of the BOUTEF corpus.}
    \label{fig:boutef_overall}
\end{figure}

Figure~\ref{fig:boutef_overall} provides a visual summary of the BOUTEF corpus structure, 
illustrating the five content types distributed across its two main components: BOUTEF-Fake, 
which encompasses fake news articles, comments on fake news, and debunking posts; and 
BOUTEF-Real, which comprises genuine and verified posts along with their associated comments.

\subsection{Languages and dialects within BOUTEF}
The analysis of the BOUTEF corpus highlights strong multilingual dynamics and rich expressive lexicons.
Modern Standard Arabic (MSA) is the dominant language in fake news posts, likely because it maximizes comprehensibility across the Arabic-speaking audience, and debunking content is also predominantly written in MSA. By contrast, user interactions rely more on dialects, giving comments a more personal and conversational tone in which idiomatic expressions and Arabizi often signal skepticism, irony, or disbelief.
French and MSA are more frequent in formal rebuttals, whereas Arabizi and dialectal variants are more common in spontaneous and emotionally charged reactions. Table \ref{Lang} summarizes the distribution of languages and dialects in the dataset.

\begin{table}[htbp]
\begin{tabular}{lllllll}
\hline
{\bf Language or} & {\bf Fake} & {\bf Fake} & {\bf NoFake} & {\bf Genuine}&{\bf Genuine}&{\bf Total} \\
{\bf Dialect} &  & {\bf Comment} &  & &{\bf Comment}& \\
\hline
ALGDIA  & 44& 1,191 & 2 &3&2,730& 3,970\\
ANG & 29 & 73 & 21 &20&4,486& 4,629\\
ARA&0&0&0&10&28,849&28,859\\
ARABIZI & 1 & 193 & 0 &9&1,778&1,981\\
FRA & 185 & 1,178 & 136 &124&11,914 &13,537\\
MAR & 6 & 42 & 0 &0&8 &56 \\
MARABIZI & 0 & 0&0&0 & 2& 2 \\
MSA & 578 & 1,366 & 280 & 124&105&2,341\\
SWITCHING & 32 & 501 & 9 & 7&22&571 \\
TUNDIALECT & 16 & 493 & 0 & 2&355&866 \\
TUNARABIZI  & 0 &89 &  0 & 10& 72&    171\\
\hline
{\bf Total}&{\bf 891}&{\bf 5,126}&{\bf 448}&{\bf 309}&{\bf 50,321}&{\bf 57,095}\\
\hline\
\end{tabular}
\caption{Distribution of languages and dialects in the BOUTEF dataset. ALGDIA: Algerian dialect, ANG: English, ARA: Arabic not identified precisely of scraped text, ARABIZI: Algerian text written in Latin script, FRA: French, MAR: Moroccan dialect, MARABIZI: Modern standard arabic written in latin script, MSA: Modern Standard Arabic, SWITCHING: Any combination of two dialects or languages, TUNDIALECT: Tunisian dialect and TUNARABIZI: Tunisian dialect written in latin script.}
\label{Lang}
\end{table}

Table \ref{tab:post_types} presents the three narrative acquisition modes used in BOUTEF. The corpus is largely dominated by \textbf{SCRAPED} narratives (50,297 items, about 88.1\%), highlighting the role of large-scale keyword-based collection for broad coverage and diversity; this category includes only genuine posts and their associated comments. \textbf{CHECKED} narratives (6,280 items, about 11.0\%) form a substantial manually verified subset that improves data reliability and provides high-confidence material for evaluation; this category includes fake news narratives, their comments, and debunking posts. By contrast, \textbf{FABRICATED} narratives (183 items, about 0.3\%), generated with few-shot LLM prompting and then manually enriched with metadata, represent a small but useful subset for stress-testing models on synthetic misinformation patterns; this category concerns fake news narratives. Overall, this distribution reflects a deliberate balance between scale (SCRAPED), quality control (CHECKED), and controlled experimentation (FABRICATED). 
\begin{table}[h]
\centering
\begin{tabular}{l|r}
\hline
\textbf{PostType} & \textbf{Count} \\
\hline
CHECKED & 6,280 \\
FABRICATED & 183 \\
SCRAPED & 50,297 \\
\hline
\end{tabular}
\caption{Distribution of the Post types.}
\label{tab:post_types}
\end{table}

\section{Data processing and analytical methodology for multilingual fake news investigation}
We adopted a mixed-methods approach designed to capture both the breadth and depth of the fake news phenomenon within the BOUTEF corpus. The first phase focused on data extraction and preprocessing to ensure dataset integrity. Standardized text processing routines and normalization techniques were applied to handle linguistic diversity, particularly code-switching between Arabic, French, and English, as well as spelling variations in Arabizi. For image analysis, metadata extraction and techniques were used to detect manipulations such as deepfakes or altered images.

After preprocessing, thematic coding was applied to categorize the articles into 67 major and minor themes. This process enabled the quantification of topic frequencies, the tracking of their evolution over time, and the identification of recurring patterns and emerging trends. This classification allowed us to cluster different types of fake news stories and to determine the most prominent and frequently recurring topics that fake news tends to focus on.

Sentiment and linguistic analysis offered valuable insights into the emotional and cultural layers of fake news. Using Natural Language Processing (NLP) tools, the study highlighted a strong predominance of negative emotions, particularly anger, frustration, and distrust, in user comments. Additionally, the analysis identified the frequent use of culturally specific expressions and language patterns that reflect particular social or regional contexts.

To study the interaction between misinformation and platform dynamics, we correlated social media engagement metrics (shares, comments) with factors such as the number of followers of fake news creators and post timing. This highlighted the amplification role of social media and influencers in spreading misinformation.

Finally, a comparative framework segmented the dataset by region (Algeria vs. Tunisia). Statistical tests and cluster analysis identified significant differences in themes, manipulation techniques, and public reactions, as well as shared trends shaped by distinct sociopolitical contexts.

This methodology provides a nuanced understanding of how fake news emerges and spreads, supporting the development of strategies for combating misinformation and promoting media literacy.

\section{Methods of manipulation and fabrication}

Iswara et al. \cite{iswara2020manipulation} showed that creators of hoaxes frequently rely on unconventional spellings, emoticons, manipulated images, and persuasive speech acts to spread misinformation. Many of these linguistic and rhetorical strategies are also present in fake news, though not all are relevant to our dataset. Essentially, both hoax and fake news creators exploit language and style to manipulate audiences.

Analysis of BOUTEF further reveals recurring mechanisms through which fake news creators craft persuasive and deceptive content. These mechanisms operate across linguistic, visual, and rhetorical levels, often working in combination to maximize impact. Specifically, linguistic and emotional alterations, including unconventional spellings (e.g., Macr0n), exaggerated punctuation (Totalitarian REGIME!!!), and alarmist emoji use (\raisebox{-0.5em}{\includegraphics[scale=0.7]{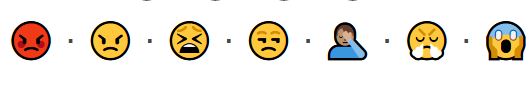}}) amplify outrage and bypass traditional filters of credibility.

Visual manipulation often involves the alteration or fabrication of images depicting political figures, such as French President Emmanuel Macron, in compromising, weakened, or isolated situations, with the apparent aim of damaging their public image and undermining their legitimacy.

Sensationalist language is common in headlines framed as shocking revelations (e.g., You will not believe what Macron signed in secret!) to create urgency and foster virality. False authority is constructed through references to non-existent institutes or experts, such as the so-called European Institute of Verification, providing a façade of legitimacy. Media imitation is observed in websites that mimic reputable outlets to spread fabricated narratives (for example, Macron prepares to abolish the right to strike), exploiting readers’ trust in established news formats.

To synthesize, these techniques align with widely recognized manipulation practices: emotional amplification, visual deception, sensational language, reference to false sources, and imitation of credible media formats. 
These mechanisms illustrate how misinformation leverages both linguistic creativity and multimodal resources to manipulate perception, trigger emotional responses, and strengthen narrative plausibility.

\subsection{Emotional language and selective truths in fake news}

Fake news creators use rhetorical strategies such as emotional language, sensational headlines, and evocative framing to manipulate readers. In BOUTEF, these strategies frequently combine emotionally charged wording with selective quoting (cherry-picking), which blends factual elements with misleading interpretation and creates an illusion of credibility. Table \ref{tab:emotional_examples} presents representative examples and their English translations.

\begin{table}[h]
\centering
\small
\begin{tabular}{|p{0.30\linewidth}|p{0.30\linewidth}|p{0.32\linewidth}|}
\hline
\textbf{Original example} & \textbf{English translation} & \textbf{Manipulative effect} \\
\hline
{\it 2 mois de pleurs. Les Algériens n'ont même plus pitié d'eux-mêmes.} & {\it Two months of tears. Algerians no longer even pity themselves.} & Amplifies despair and collective humiliation after a sports defeat. \\
\hline
{\it Record du plus long deuil dans l'histoire d'un pays.} & {\it Record of the longest mourning period in a country's history.} & Uses hyperbole to dramatize national shame and intensify emotional reaction. \\
\hline
{\it Le régime de Tebboune manipule outrageusement des chiffres ... déclin alarmant des volumes de nos hydrocarbures.} & {\it Tebboune's regime outrageously manipulates figures ... alarming decline in our hydrocarbon volumes.} & Cherry-picks economic claims and relies on loaded terms (e.g., {\it outrageously}, {\it alarming}) to trigger anger while appearing factual. \\
\hline
{\it Match à rejouer c'est acceptable, le rejouer en Hongrie c'est de la rigolade, que des mensonges bien décorés.} & {\it Replaying the match is acceptable; replaying it in Hungary is ridiculous, just well-decorated lies.} & Mixes a real event with corruption accusations to delegitimize arbitration decisions. \\
\hline
\end{tabular}
\caption{Examples of emotional and selective-truth rhetoric in BOUTEF fake news, with English translations.}
\label{tab:emotional_examples}
\end{table}

These examples show how fear, anger, and humiliation are strategically activated while preserving a surface form that mimics credible reporting.

\subsection{Visuals as weapons in spreading false information}

Visual media is a powerful tool for fake news, with over 22\% of articles including images or videos in BOUTEF. These range from genuine photos taken out of context to heavily manipulated graphics. Forensic analysis reveals common practices like cropping, color adjustment, and misleading captions \cite{cao2020exploring}. 

\begin{figure}[!h]
\centering
\includegraphics[scale=0.7]{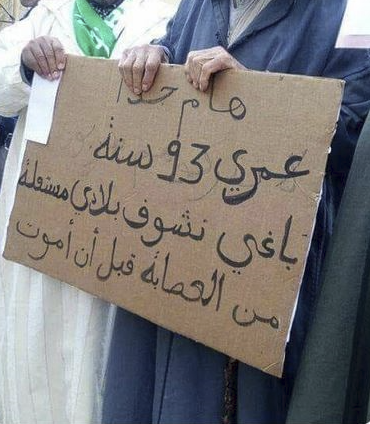}
\caption{An altered image depicts a person stating, {\small At 93 years old, I would like to see my country free from the gang before I die.} "gang" is used metaphorically to refer to the old government.}
\label{Vieux}
\end{figure}

Figure \ref{Vieux} illustrates a case where imagery and text combine to deceive. A photo of two men holding a protest sign is matched with a caption about poverty in Algeria. However, their clothing and phrasing indicate they are not Algerian, fabricating a misleading narrative of exploitation. Visual literacy alongside local social and cultural knowledge is thus essential for detecting such manipulations.

\subsection{Political manipulation}

Fake news is often used to manipulate political discourse in Algeria and Tunisia. Articles target opponents, influence elections, or foster distrust in institutions. Strategic messaging relies on emotional appeals and selective facts to resonate with existing anxieties and biases. Surveys from the BOUTEF corpus show that exposure to such content correlates with declining trust in media and government, thereby undermining democratic processes. This erosion of credibility also fuels social fragmentation, reinforcing echo chambers and polarizing communities.
The analysis of the BOUTEF corpus reveals that fake news narratives in Algeria and Tunisia rely on recurring rhetorical and strategic patterns. These include emotional framing, political targeting, selective use of facts, and attempts to undermine trust in institutions. Table~\ref{tab:political_misinfo_examples} presents representative examples extracted from the corpus, alongside their English translations and the corresponding misinformation strategies.

\begin{table}[h]
\centering
\small
\begin{tabular}{|p{0.28\linewidth}|p{0.30\linewidth}|p{0.34\linewidth}|}
\hline
\textbf{Arabic Narrative} & \textbf{English Translation} & \textbf{Misinformation Strategy} \\
\hline
\RL{تونس تحت الاحتلال} & Tunisia under occupation & Emotional and sensational framing; fear amplification; misleading context. \\
\hline
\RL{تونس أصبحت مملكة للأفارقة} & Tunisia has become the kingdom of Africans & Xenophobic narrative; polarization; exploitation of social anxieties. \\
\hline
\RL{حسب إحصائيات رسمية، عدد المهاجرين في ارتفاع خطير} & According to official statistics, the number of migrants is dangerously increasing & Cherry-picking/false attribution; illusion of legitimacy; misuse of data. \\
\hline
\RL{استطلاع رأي: نسبة التأييد للرئيس مرتفعة بشكل غير مسبوق} & Poll: support for the president is at an unprecedented level & Fabricated data; manipulation of public opinion; electoral influence. \\
\hline
\RL{الشعب يعيش حالة من الغضب والخوف بسبب الأوضاع} & The people are living in a state of anger and fear due to the situation & Emotional amplification; fear-based narrative; cognitive manipulation. \\
\hline
\end{tabular}
\caption{Examples of misinformation narratives and strategies in the political domain. Note that translations are contextual rather than literal, preserving the intended meaning over word-for-word accuracy.}
\label{tab:political_misinfo_examples}
\end{table}

\subsection{Duration of the fake news}

\begin{figure}[!h]
\includegraphics[scale=0.45]{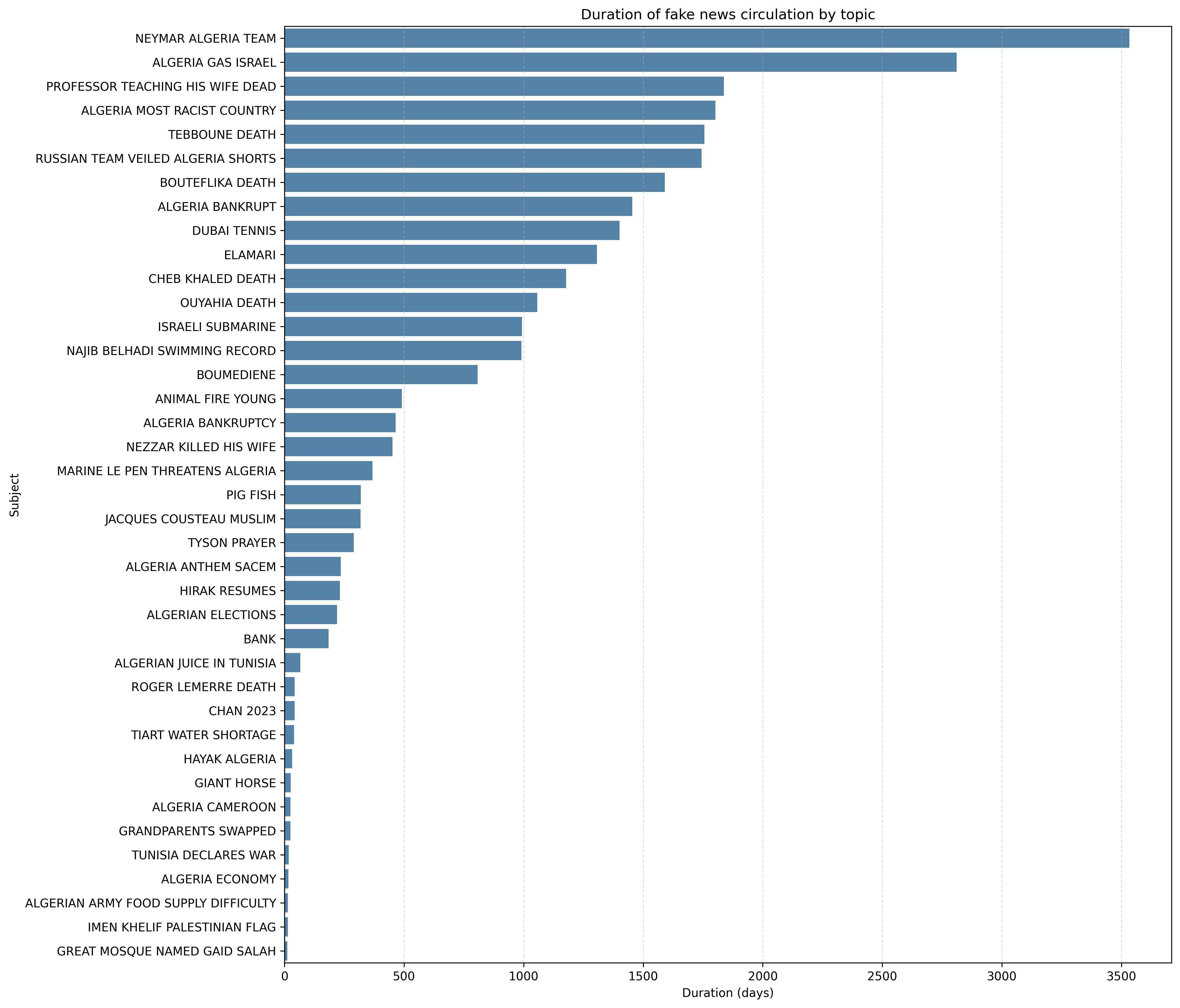}
\caption{Duration of some fake news in BOUTEF.}
\label{dur}
\end{figure}

Misinformation can persist for years, even after debunking. Prior studies \cite{Gwon2024,Manolis2021} show that timely interventions reduce lifespan, but our findings indicate resilience in certain narratives. Figure \ref{dur} illustrates the lifespan of selected cases in BOUTEF.

\begin{figure}[!h]
\centering
    \includegraphics[scale=0.22]{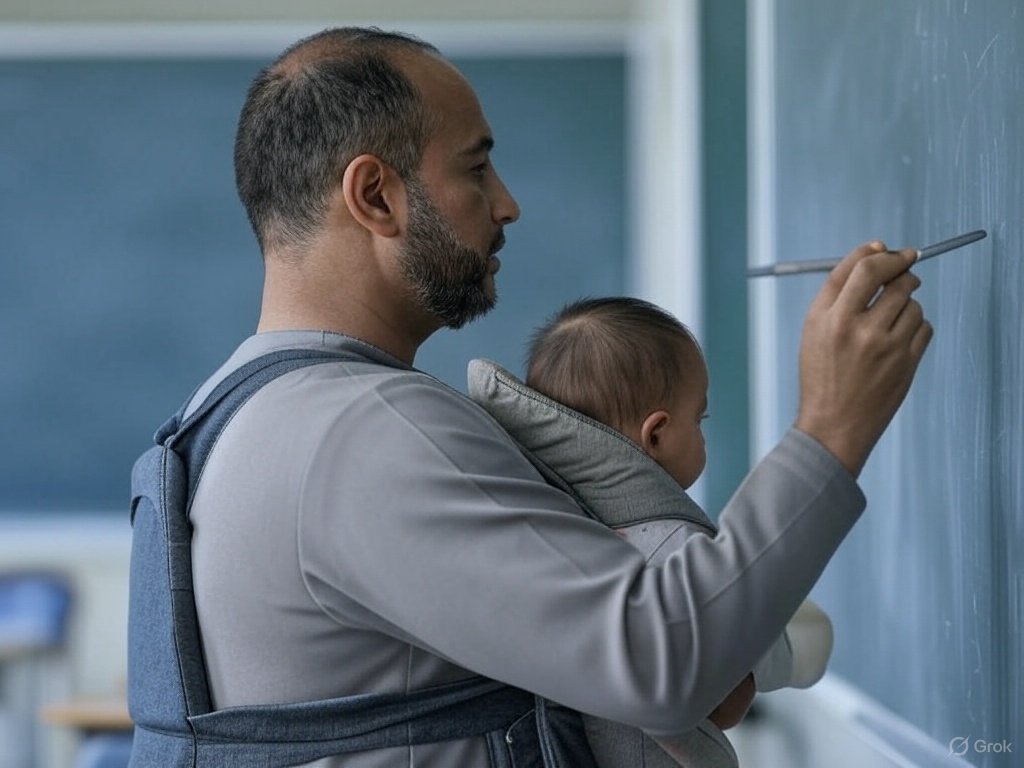}
    \caption{Misinformation concerning the state of a professor}
    \label{Prof}
\end{figure}

For instance, the false claim that Neymar was of Algerian origin circulated for 3,533 days (9 years), while the rumor of President Tebboune’s death persisted for 1,756 days (4 years). Even quickly debunked stories, such as the widely shared narrative about a professor who became father after his wife’s death, remained online for 1,837 days (5 years) (Figure \ref{Prof}). 

The persistence of these emotionally charged stories shows that debunking alone is insufficient. Their strong emotional appeal, empathy, admiration, or outrage, makes them memorable and shareable, sustaining their circulation across countries and years. In some cases, misinformation may even disappear completely for a period and then re-emerge later when a favorable social, political, or emotional context revives its visibility. Fact-checking is often ineffective due to psychological and sociological barriers. Confirmation bias leads individuals to reject corrections that challenge their beliefs, while the backfire effect can reinforce false narratives~\cite{Nyhan2010}. Sociologically, echo chambers and social identity dynamics amplify misinformation within polarized groups, as seen in the BOUTEF corpus with politically charged disinformation. The rapid spread of misinformation on platforms like Twitter/X outpaces the slower fact-checking process \cite{Vosoughi2018}.

\section{Topic frequency and themes}

Zhou et al. \cite{zhou2023does} found that across languages, fake news tends to exhibit a broader range of topics and greater uncertainty compared to real news. Building on this, we study the topics and forms of fake news in the BOUTEF corpus.

\subsection{Quantitative analysis of message length: statistical insights}

Table \ref{tab:stats_type_typemessage} reveals clear differences in message length across sources and message types. Within the \textbf{CHECKED} subset, \texttt{NoFake} items are by far the longest (mean=70.75, median=58.5), whereas \texttt{FakeComment} texts are much shorter (mean=13.89, median=9), which is consistent with concise and reactive user behavior. The relatively large standard deviations and high maxima in several checked categories (e.g., max=735 for \texttt{NoFake}, max=594 for \texttt{Fake}) indicate strong heterogeneity and the presence of long-tail cases. By contrast, \textbf{FABRICATED} messages are short and tightly distributed (\(\sigma\) around 3.2--3.4), which is consistent with their construction through few-shot learning prompts rather than naturally occurring discourse. Finally, \textbf{SCRAPED} \texttt{GenuineComment} dominates in volume (N=50,226) but remains short on average (mean=19.19, median=12), confirming that large-scale collected engagement is mostly made of brief comments, with a few extreme outliers (max=1160).

\begin{table}[htbp]
\centering
\small
\begin{tabular}{llrrrrrr}
\toprule
\textbf{TYPE} & \textbf{TYPE\_MESSAGE} & \textbf{N} & \textbf{Mean} & \textbf{Min} & \textbf{Max} & \textbf{$\sigma$} & \textbf{Median} \\
\midrule

\multirow{5}{*}{\textbf{CHECKED}} 
& Fake             & 708  & 26.43 & 2   & 594 & 37.63 & 17.0  \\
& FakeComment      & 5,126 & 13.89 & 1   & 397 & 16.31 & 9.0   \\
& NoFake           & 448  & 70.75 & 2   & 735 & 60.68 & 58.5  \\
& GENUINE          & 226  & 28.38 & 4   & 112 & 15.40 & 28.0  \\
& GenuineComment   & 95   & 21.75 & 1   & 327 & 39.30 & 13.0  \\
\midrule

\multirow{2}{*}{\textbf{FABRICATED}} 
& Fake             & 183  & 16.52 & 10  & 27  & 3.39  & 16.0  \\
& GENUINE          & 12   & 10.67 & 7   & 16  & 3.17  & 11.0  \\
\midrule

\multirow{2}{*}{\textbf{SCRAPED}} 
& GeniuneComment   & 50,226 & 19.19 & 1    & 1,160 & 28.44 & 12.0  \\
& GENUINE          & 71    & 11.10 & 3    & 67   & 7.68  & 10.0  \\
\bottomrule
\end{tabular}
\caption{Distribution of narratives according to their category.}
\label{tab:stats_type_typemessage}
\end{table}

"Fake" messages show substantial variability, with lengths ranging up to 594 words and a distribution skewed toward shorter texts. "FakeComment" posts are generally shorter and more consistent, while "NoFake" messages are much longer on average (mean = 70.75 words) and highly variable, reflecting more detailed legitimate discourse. These differences suggest that message length alone cannot reliably distinguish categories, but it highlights contrasting communicative styles.

\begin{figure}[!h]
\includegraphics[scale=0.5]{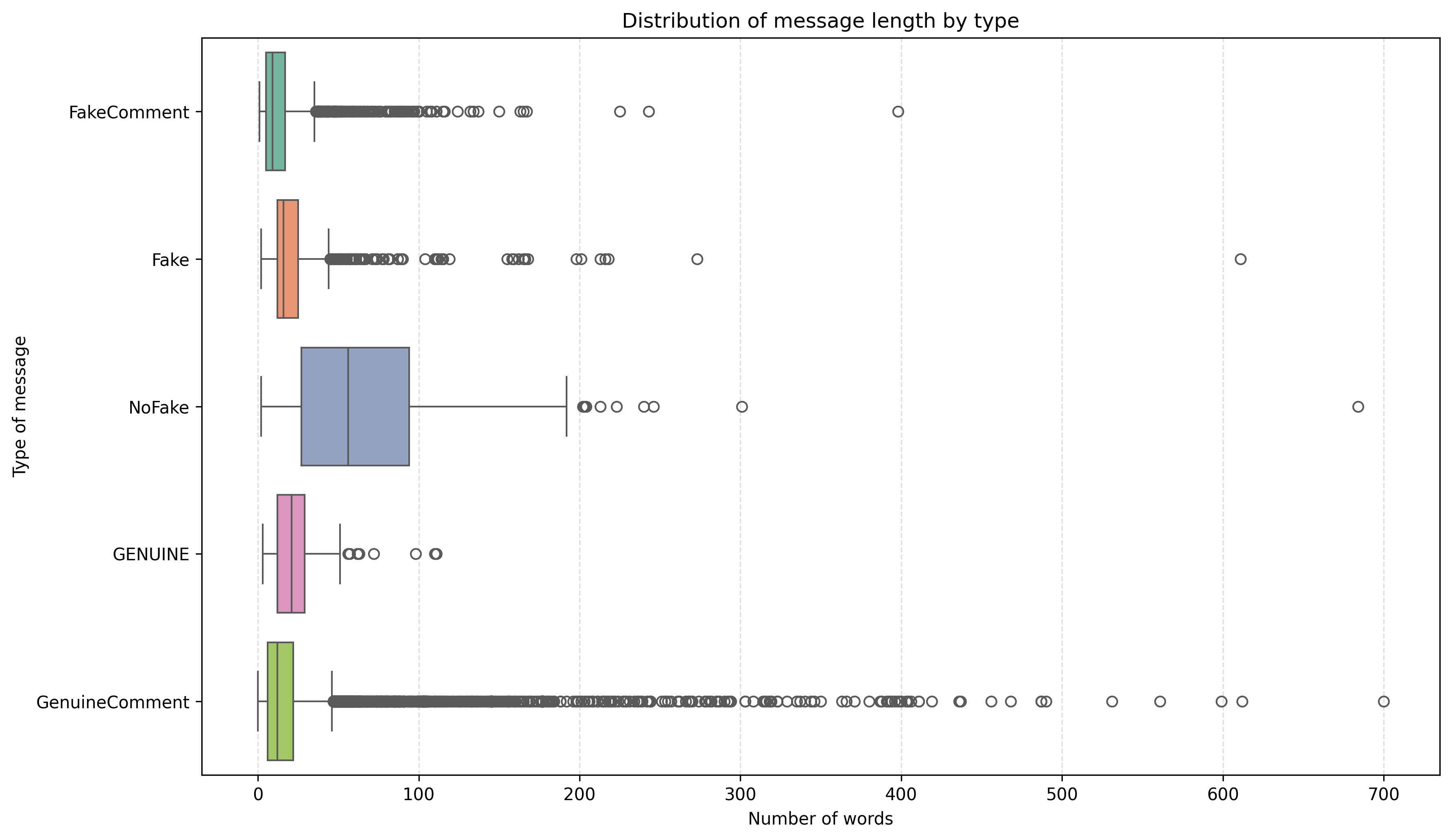}
\caption{Distribution of message length by type.}
\label{Moustache}
\end{figure}

The boxplot of Figure \ref{Moustache} summarizes the distribution of message length across the different message types. The central line inside each box corresponds to the median, or Q2, which indicates the typical message length for each category. The lower and upper bounds of the box represent Q1 and Q3, respectively, and together they define the interquartile range, which contains the middle 50\% of the observations. The NoFake category has the highest median and the widest interquartile range, showing that these messages are generally longer and more variable than the others, which may reflect the explanatory nature of debunking messages. In contrast, FakeComment and Fake messages have lower medians and narrower boxes, which suggests shorter and more concentrated distributions. The GENUINE category also exhibits a relatively compact distribution, with a median close to the center of a small interquartile range. The GenuineComment group has a low median but a long right tail, indicating the presence of many short messages and a few exceptionally long ones. The numerous points beyond the whiskers represent outliers, meaning messages that are unusually long compared with the rest of their group. These outliers are especially visible in GenuineComment and NoFake, which suggests a heavy-tailed distribution. Overall, the boxplot shows that message length varies substantially across categories and may serve as a useful discriminative feature, even though it does not fully separate the groups.

\subsection{Topical analysis}

The BOUTEF corpus contains 67 identified themes (Appendix A). Political narratives dominate, but the thematic distribution differs between Algeria and Tunisia.

\begin{table}[htbp]
\centering
\begin{tabular}{lcc}
\toprule
\textbf{Major theme} & \textbf{Fake (\%)} & \textbf{Genuine (\%)} \\
\midrule
Politics          & 18.03 & 12.58 \\
Sports              & 8.78 & 13.25 \\
Religion           & 5.80  & 1.32  \\
Death personnality&5.17&1.32\\
Army              & 5.17  & 2.65  \\
Economy           & 3.45   & 24.50 \\
Culture            & 2.66  & 8.61\\
Denigration &3.29&--\\
Espionnage&    3.29&1.32\\

\midrule
\textbf{Other themes} & 49.53 & 34.45 \\
\bottomrule
\end{tabular}
\caption{Top 10 major themes in Algeria – distribution in percentages. Percentages are computed over the total number of entries in the BOUTEF corpus, across both Fake and Genuine content types.}
\label{tab:themes_algerie_fake_genuine}
\end{table}

For the Algerian part of BOUTEF (Table \ref{tab:themes_algerie_fake_genuine}), \textbf{Politics} is the leading theme in both fake and genuine narratives, but it is more prominent in fake content (18.03\%) than in genuine content (12.58\%). Beyond politics, the two distributions diverge clearly. Genuine narratives give much more space to \textbf{Economy} (24.50\%), \textbf{Sports} (13.25\%), and \textbf{Culture} (8.61\%), suggesting a stronger focus on public information, socioeconomic issues, and everyday news. By contrast, fake narratives place greater emphasis on \textbf{Religion} (5.80\%), \textbf{Death personality} (5.17\%), and \textbf{Army} (5.17\%), all of which are themes likely to trigger emotion, fear, or symbolic reactions. In addition, \textbf{Denigration} appears among the main fake themes but not among the major genuine ones, reinforcing the idea that fake narratives in Algeria rely more heavily on polarizing or reputational content. The \textbf{Other themes} category is also much larger for fake content (49.53\%) than for genuine content (34.45\%), which indicates a broader thematic dispersion in misinformation than in verified reporting.

\begin{table}[htbp]
\centering
\small
\begin{tabular}{lcc}
\toprule
\textbf{Major Theme} & \textbf{Fake (\%)} & \textbf{Genuine (\%)} \\
\midrule
War                    & 14.46 & --     \\
Economy                 & 14.46 & 32.18  \\
Politics               &9.92  & 9.20   \\
Immigration            & 9.50  & 8.05     \\
Sport                  &4.96 & 10.34  \\
Health                 & 2.89  & 11.49  \\
Disaster               & 2.89  & --     \\
Tourism&2.89&--\\
Culture                & 1.65  & 11.49  \\
\midrule
\textbf{Other themes}  & 36.06 & 17.25  \\
\bottomrule
\end{tabular}
\caption{Distribution of major themes in Tunisia according to content type (in percentages).}
\label{tab:tunisia_major_themes}
\end{table}

For the Tunisian part of BOUTEF (Table \ref{tab:tunisia_major_themes}), the contrast is even more pronounced. In fake narratives, \textbf{War} and \textbf{Economy} are tied as the main themes (14.46\% each), followed by \textbf{Politics} (9.92\%) and \textbf{Immigration} (9.50\%), revealing a concentration on conflict, instability, and anxiety-related topics. In genuine narratives, \textbf{Economy} clearly dominates (32.18\%), while \textbf{Health} and \textbf{Culture} each reach 11.49\%, \textbf{Sport} accounts for 10.34\%, and \textbf{Politics} remains comparatively stable at 9.20\%. Some themes, such as \textbf{War}, \textbf{Disaster}, and \textbf{Tourism}, appear among the major fake themes but not among the major genuine ones, which suggests that fake content in Tunisia relies more strongly on exceptional, crisis-oriented, or sensational topics. As in Algeria, the \textbf{Other themes} category is substantially larger in fake narratives (36.06\%) than in genuine ones (17.25\%), confirming that misinformation tends to be more thematically scattered, whereas genuine content is more structured around recurring public-information domains.

\subsection{Dependency of message type according to theme}

In the following, the chi-square ($\chi^2$) test of independence is employed in this study to examine whether a statistically significant relationship exists between the major theme of a message and its associated label, namely fake or genuine. This test is particularly appropriate  as it evaluates whether the observed distribution of frequencies across thematic categories deviates from what would be expected under the assumption of independence between variables. In other words, it allows us to determine whether certain themes are disproportionately associated with either fake or genuine messages.

\begin{table}[htbp]
\centering
\begin{tabular}{ll}
\toprule
\textbf{Statistic} & \textbf{Value} \\
\midrule
$\chi^2$ & 272.99 \\
Degrees of freedom & 34 \\
p-value & $4.12 \times 10^{-39}$ \\
Interpretation & Highly significant \\
Cram\'er's V & 0.49 \\
\bottomrule
\end{tabular}
\caption{Chi-square test and effect size results.}
\label{tab:chi_square_results}
\end{table}
The chi-square test indicates a statistically significant association between the major theme of a message and its label, namely \textit{fake} or \textit{genuine}. For this analysis, we retained only the \texttt{MAJOR\_THEME} categories whose frequencies were greater than 7, in order to avoid unstable conclusions driven by very rare themes. The obtained result, with $\chi^2 = 272.99$, $34$ degrees of freedom, and a p-value close to zero, leads to the rejection of the null hypothesis of independence between \texttt{MAJOR\_THEME} and \texttt{TYPE\_MESSAGE}. Moreover, Cram\'er's $V = 0.49$ indicates a moderately strong to strong association, showing that the relationship is not only statistically significant but also meaningful in practical terms. This suggests that the distribution of themes differs substantially between genuine and fake messages.

\begin{figure}[H]
    \centering
    \includegraphics[width=\linewidth]{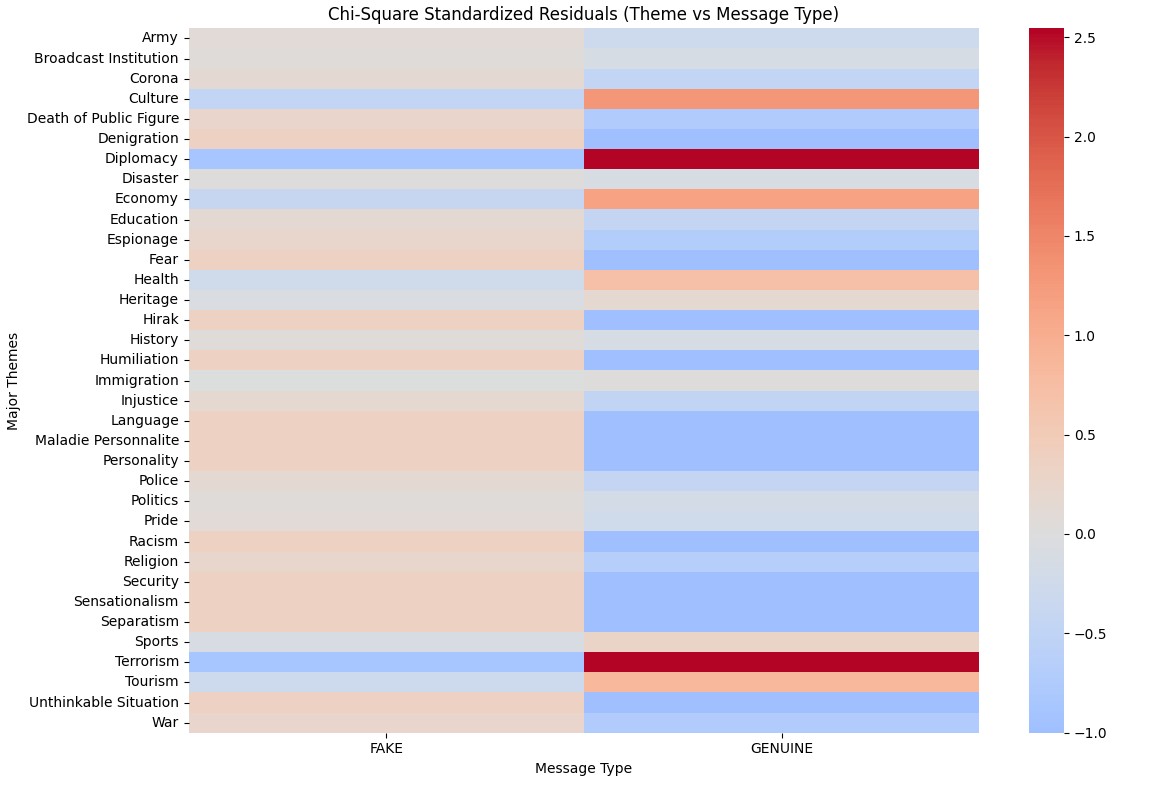}
    \caption{Chi-square standardized residuals for theme and message type.}
    \label{Heat}
\end{figure}

The residual heatmap given in Figure \ref{Heat} provides a more detailed view of this dependency by highlighting the deviations between observed and expected frequencies under the assumption of independence. Red cells indicate themes that are overrepresented in a given class, whereas blue cells indicate themes that are underrepresented. The stronger the color intensity, the larger the deviation from expectation.

Overall, Figure \ref{Heat} suggests that several themes are more strongly associated with genuine messages, including \textit{Diplomacy}, \textit{Terrorism}, \textit{Economy}, \textit{Culture}, \textit{Health}, and \textit{Tourism}. Conversely, other themes appear to be more closely related to fake messages, such as \textit{Denigration}, \textit{Humiliation}, \textit{Personality}, \textit{Racism}, \textit{Language}, and \textit{Personal Illness}. These patterns indicate that thematic content contains informative cues that may help distinguish between genuine and fake messages.

Nevertheless, this result should be interpreted as evidence of association rather than causality. A theme being more frequent in fake messages does not imply that the theme itself generates falsity; it only shows that some themes occur disproportionately in one category. For instance, \textit{Diplomacy} may be more prevalent in genuine messages, whereas \textit{Denigration} may appear more often in fake ones. Such differences are particularly relevant for exploratory analysis and may also support feature engineering for downstream classification tasks.
\subsection{Distribution of the fakes category}
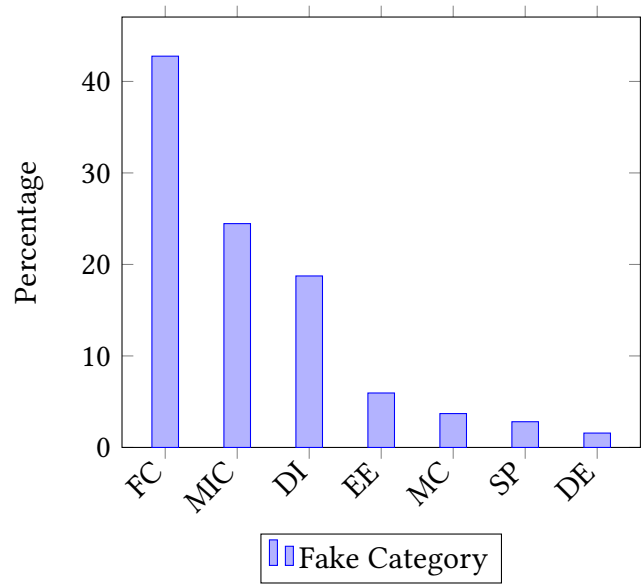
\begin{figure}[!h]
\begin{tikzpicture}
    \begin{axis}[
        ybar,
        ymin=0,
        ylabel=Percentage,
        symbolic x coords={FC,MIC, DI, EE, MC, SP, DE},
        xtick=data,
        x tick label style={rotate=45,anchor=east},
        legend style={at={(0.5,-0.2)},anchor=north,legend columns=-1}
    ]
    \addplot coordinates {
        (FC, 42.76)
        (MIC, 24.46)
        (DI, 18.74)
        (EE, 5.95)
        (MC, 3.70)
        (SP, 2.81)
        (DE, 1.57)
    };
    \legend{Fake Category, Total}
    \end{axis}
\end{tikzpicture}
\caption{Percentage distribution of fake news categories in BOUTEF (FC: Fabricated Content; MIC: Misleading Content; DI: Diverted Information; EE: Exaggeration or Extension; MC: Manipulated Content; SP: Satire or Parody; DE: Deny the Existence).}
\label{Cat}
\end{figure}

In Figure \ref{Cat}, we are interested by the distribution of fake news in accordance to Wardle's categories.
The percentage  indicates a strongly asymmetric structure of misinformation strategies. \textbf{FC} (42.76\%) is by far the most frequent category, showing that fully fabricated narratives constitute the core of fake content in the corpus. The second and third categories, \textbf{MIC} (24.46\%) and \textbf{DI} (18.74\%), remain substantial and together with \textbf{FC} account for 85.96\% of all cases, which suggests that most misinformation relies either on complete fabrication, misleading framing, or diversion of otherwise factual information. In contrast, \textbf{EE} (5.95\%), \textbf{MC} (3.70\%), \textbf{SP} (2.81\%), and \textbf{DE} (1.57\%) form a relatively small tail. Their lower prevalence indicates that exaggeration, content manipulation, satirical/parodic framing, and denial narratives are present but play a secondary role compared with the three dominant categories.

\subsection{Demographic factors and gender analysis}

Demographic signals, although limited, suggest a diversity of creator profiles. In the broader literature, younger authors are often associated with Arabizi and more digital writing styles, whereas older authors tend to rely on more traditional rhetoric. Gender analysis adds further nuance. However, this remains a general observation that cannot be verified within BOUTEF, since the corpus does not provide information about users' ages.
Almenar et al. \cite{almenar2021gender} studied gender differences in fake news perception. The article finds that women are more concerned than men about the spread of fake news, but both genders face the same difficulty in detecting it. It also shows that fake news is mainly associated with politics, and this is true for both men and women.
Overall, the study suggests that gender differences are small in fake-news reception, even though women report higher levels of concern. The authors conclude that disinformation is a widespread problem that affects everyone similarly in terms of detection, while gender mainly shapes the level of concern.
In BOUTEF, linguistic analysis shows that men’s contributions are often more assertive, while women’s posts display relational language. Representation is uneven, with women underrepresented.

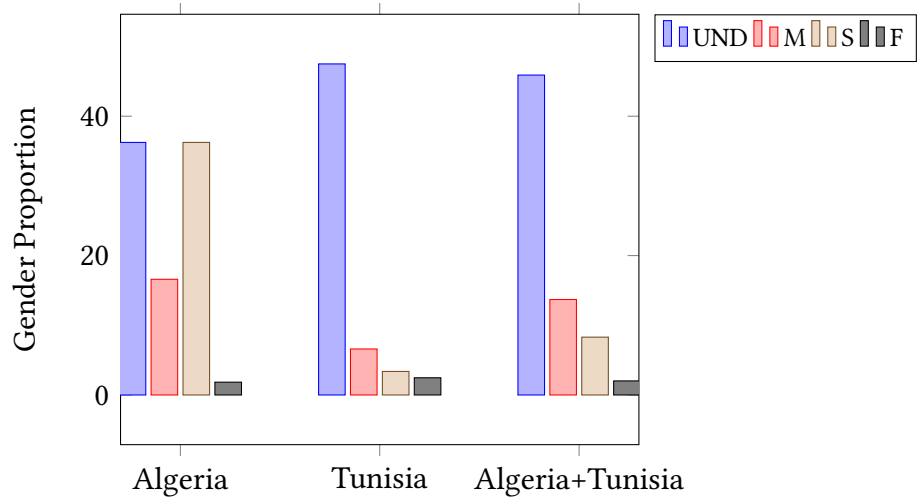
\begin{figure}[H]
\begin{tikzpicture}
    \begin{axis}[
        ybar,
        ymin=0,
        ylabel=Gender Proportion,
        symbolic x coords={Algeria, Tunisia, Algeria+Tunisia},
        xtick=data,
        legend pos=outer north east,
        legend columns=4,
        legend style={font=\footnotesize},
        enlargelimits=0.15
    ]
    \addplot coordinates {(Algeria, 36.24) (Tunisia, 47.5) (Algeria+Tunisia, 45.9)}; 
    \addplot coordinates {(Algeria, 16.6) (Tunisia, 6.6) (Algeria+Tunisia, 13.7)}; 
    \addplot coordinates {(Algeria, 36.24) (Tunisia, 3.38) (Algeria+Tunisia, 8.3)}; 
    \addplot coordinates {(Algeria, 1.84) (Tunisia, 2.47) (Algeria+Tunisia, 2.02)}; 
    \legend{UND, M, S, F}
    \end{axis}
\end{tikzpicture}
\caption{Distribution of gender in BOUTEF (UND: undetermined gender; S: website without an author; M: male; F: female).}
\label{genre}
\end{figure}

Thematic preferences also differ by gender. Table \ref{ThemeFM} summarizes the main fake-news themes associated with male and female users.

\begin{table}[H]
\centering
\begin{tabular}{lrr}
\hline
\textbf{MAJOR\_THEME} & \textbf{M (\%)} & \textbf{F (\%)} \\ \hline
POLITICS             & 19.13 & 11.76 \\ 
SPORTS                &8.69 & 0\\
IMMIGRATION &0&11.76\\
WAR                  & 7.82 & 5.88 \\ 
ARMY  &6.95&5.88\\
ECONOMY&6.08&5.88\\
 \hline
\end{tabular}
\caption{Distribution of fake news major themes for men (M) and women (F).}
\label{ThemeFM}
\end{table}

This table should be interpreted together with Figure \ref{genre}, which shows that a large share of fake news is spread by \textbf{UND} users, that is, profiles that could not be reliably classified as men, women, or websites. The male/female percentages therefore describe only the identifiable subset of users. Within that subset, men are more associated with \textbf{Politics} (19.13\%), \textbf{Sports} (8.69\%), and \textbf{War} (7.82\%), whereas women are more represented in \textbf{Immigration} (11.76\%). \textbf{Politics} remains important for both groups, but it is more prevalent among men than among women (11.76\%). Overall, the results suggest some topical differences by gender, while also highlighting the uncertainty created by the large proportion of undetermined profiles.

\subsection{Public reactions and engagement}
Public reactions range from endorsement to skepticism and debunking. Tandoc et al. \cite{Tandoc02072024} showed that terminology shapes responses to misinformation. In BOUTEF, as we show below by using word clouds, sensationalist posts attract emotionally charged comments, often reinforcing narratives. 
\subsubsection{Analysis of the most frequent words in BOUTEF-Fake}
\begin{figure}[H]
\includegraphics[scale=0.35]{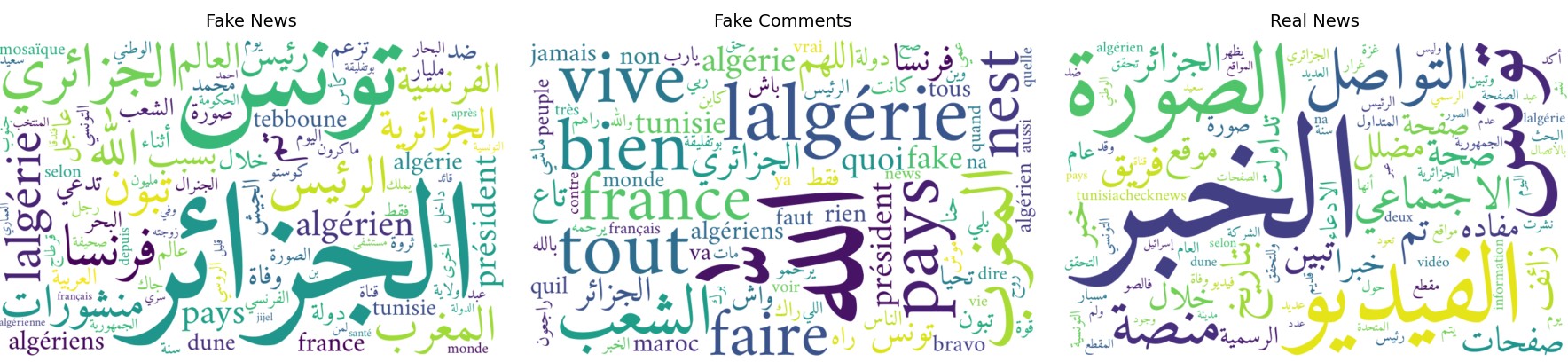}
\caption{Word cloud of fake.}
\label{CloudeFake}
\end{figure}

Word clouds reveal clear lexical contrasts across fake posts, fake comments, and debunking content (Figure \ref{CloudeFake}).
 In fake posts, the most frequent terms are geopolitical and identity-focused, including \RL{الجزائر} ({\it Algeria}), \RL{تونس} ({\it Tunisia}), \RL{الجزائري} ({\it Algerian}), and references to France (e.g., \RL{فرنسا}, \textit{france}). Political markers are also salient, such as \RL{الرئيس} ({\it President}), \RL{تبون}, and \textit{tebboune}. The recurrence of \RL{وفاة} ({\it death}) and \RL{عاجل} ({\it urgent}) indicates frequent use of mortality and urgency framing to maximize attention.

Fake-news comments are even more affective and reactive: \RL{الله} ({\it Allah}) is dominant (716 occurrences), followed by \RL{الجزائر} ({\it Algeria}), \RL{ربي} ({\it my God}), and \RL{مات} ({\it died}). This pattern suggests that comments intensify moral and emotional engagement around religion, identity, and loss, while maintaining a transnational focus through references to Algeria, Tunisia, Morocco, and France.
By contrast, NoFake/debunking posts are organized around a verification-oriented lexicon: \RL{الخبر} ({\it news}), \RL{الفيديو} ({\it video}), \RL{الصورة} ({\it image}), \RL{التواصل} ({\it network/communication}), \RL{منصة} ({\it platform}), \RL{الاجتماعي} ({\it social}), and labels such as \RL{مضلل} ({\it misleading}), \RL{زائف} ({\it fake}), and \RL{تبين} ({\it shown/verified}). This vocabulary reflects an evidential discourse centered on source checking, media artifacts, publication timing, and institutional validation rather than emotional amplification. 
\subsubsection{Analysis of the most frequent words in BOUTEF-Real}
The genuine part of BOUTEF shows a lexical profile that combines institutional and cross-border references with strong community engagement. In genuine posts, the most recurrent terms include \RL{تونس} ({\it Tunisia}), \RL{الجزائر} ({\it Algeria}), \textit{tunisie} ({\it Tunisia}), \textit{algérie} ({\it Algeria}), and \textit{tebboune} ({\it Algerian President}), indicating a persistent focus on Tunisia--Algeria relations, national identity, and public leadership. Additional terms such as \textit{visite} ({\it visit}), \RL{الوطني} ({\it national}), \textit{diplomatic}, sécurité (\textit{security}), and croissance (\textit{growth}) suggest that genuine narratives are more oriented toward official communication, governance, diplomacy, and socioeconomic reporting than toward alarmist framing.
\begin{figure}[!h]
\includegraphics[scale=0.5]{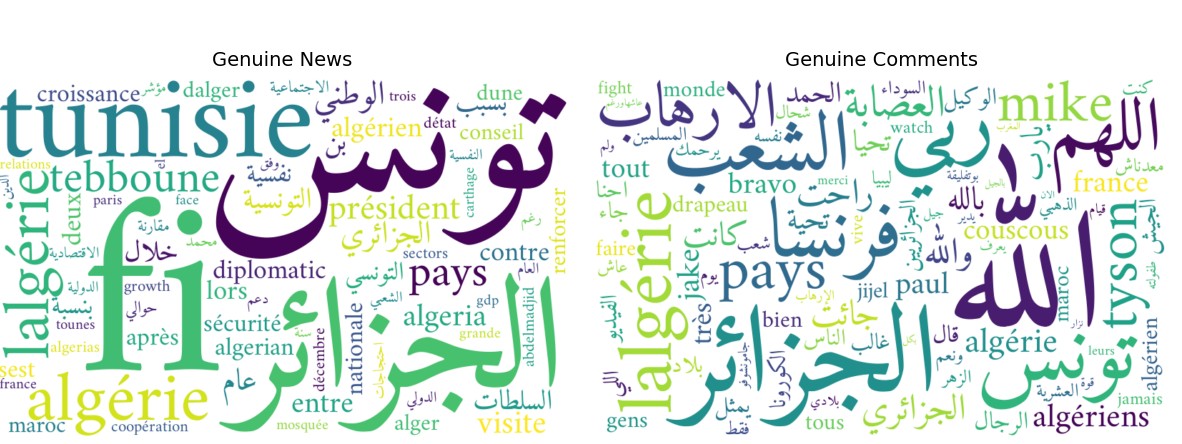}
\caption{Word cloud of genuine posts.}
\label{CloudeGen}
\end{figure}
Genuine comments, by contrast, are much more affective and participatory. Very high frequencies of \RL{الله} ({\it Allah}) which occurs 11,134 times, \RL{الجزائر} ({\it Algeria}), \RL{ربي} ({\it my God}), \RL{تونس} ({\it Tunisia}), and \RL{الشعب} ({\it the people}) show that religious expressions, collective identity, and civic sentiment dominate public reaction. The presence of terms linked to memory and crisis (e.g., \RL{الارهاب} ({\it terrorism}), \RL{العشرية} ({\it decade}), \RL{السوداء} ({\it black})) alongside supportive expressions (e.g., \RL{تحيا} ({\it long live}), \RL{تحية} ({\it salute}), bravo (\textit{congrat})) indicates that users interpret genuine information through both emotional solidarity and historical-political recall. Overall, while genuine posts are relatively institutional and informational, genuine comments reveal a socially and emotionally charged reception layer.

\subsection{Sentiment analysis}

To conduct sentiment analysis of the corpus, we use representative trigrams as lexical indicators of emotional stance in each message type. Table \ref{Trigram} presents the most recurrent trigrams in \textbf{Fake}, \textbf{FakeComment}, and \textbf{NoFake} content.

\begin{table}[h]
\centering
\scriptsize
\begin{tabular}{|p{4cm}|p{4cm}|p{4cm}|}
\hline
\textbf{Fake} & \textbf{FakeComment} & \textbf{NoFake} \\
\hline
\RL{توفت زوجته أثناء} & \RL{(فييييييييييييييييييييييييييييق ياااااااااااااااااا شعب تونس} & \RL{لتحقق تبين لخبر} \\
\RL{زوجته أثناء الولادة} & \RL{حول قوة بالله} & \RL{تبين الخبر زائف} \\
\RL{(يملك 26 فندق)} & \RL{هاو هاو هاو} & Falso \RL{فريق منصة} \\
\RL{الفيفا تقرر استبعاد} & \vspace{0.2cm}vive l'algérie vive & \RL{صفحات التواصل الاجتماعي} \\
\hline
\end{tabular}
\caption{Most recurrent trigrams in each message type.}
\label{Trigram}
\end{table}

In fake narratives, trigrams such as \RL{توفت زوجته أثناء} ({\it His wife died during}) and \RL{زوجته أثناء الولادة} ({\it his wife during childbirth}) are associated with tragedy and emotional shock, while \RL{يملك 26 فندق} ({\it he owns 26 hotels}) and \RL{الفيفا تقرر استبعاد} ({\it FIFA decides to discard}) reflect sensational and authority-based framing. In comments, expressions like \RL{حول قوة بالله} ({\it There is no power nor strength except through God.}), \RL{هاو هاو هاو} ({\it barking dog}), and \textit{vive l'algérie vive} ({\it Long live Algeria, long live}) reveal strong affective reactions ranging from faith and irony to nationalist mobilization. By contrast, NoFake trigrams (e.g., \RL{تبين الخبر زائف} ({\it the news turned out to be false}), \RL{لتحقق تبين لخبر} ({\it after verification, it appeared that the news}), \RL{صفحات التواصل الاجتماعي} ({\it social media pages})) are dominated by verification and correction cues, which correspond to a more factual and procedural register.

Taken together, these examples support sentiment-oriented interpretation of the corpus: fake content and its comments concentrate emotionally charged language. 

To obtain objective analyses, we conducted experiments on sentiment polarity across fake-news content in the full BOUTEF corpus, using well-established transformer models adapted to each language. For English, we used SieBERT \cite{hartmann2023}, a fine-tuned checkpoint of RoBERTa-large \cite{liu2019}. For French, we used multilingual BERT (mBERT)\footnote{\url{nlptown/bert-base-multilingual-uncased-sentiment}} from Hugging Face for multilingual sentiment classification. For Arabic, we used a BERT model trained for sentiment analysis in both Modern Standard Arabic and dialectal Arabic, from the CAMeL Lab suite \cite{obeid2020}.

\begin{table}[htbp]
\centering
\begin{tabular}{lr}
\toprule
\textbf{Sentiment} & \textbf{Percentage} \\
\midrule
Negative           & \textbf{59.48\%} \\
Neutral            & 30.42\% \\
Positive           & 10.10\% \\
\midrule
\textbf{Total}     & \textbf{100.00\%} \\
\bottomrule
\end{tabular}
\caption{Sentiment Distribution of Fake News Corpus.}
\label{tab:sentiment_percentage}
\end{table}

Table \ref{tab:sentiment_percentage} shows a clear dominance of negative sentiment (59.48\%), almost twice the neutral share (30.42\%) and nearly six times the positive share (10.10\%). This distribution indicates that fake-news content in the corpus is primarily framed through fear, anger, distrust, and conflict-oriented narratives. The relatively large neutral component suggests that a substantial portion of messages use an informational or pseudo-factual tone, which may increase credibility while still conveying misleading claims. By contrast, the limited positive proportion indicates that fake news in this dataset relies much less on positive persuasion than on emotionally adverse framing. 

As discussed above, some fake-news narratives aim to evoke national pride, as in the claim about Neymar falsely presenting him as originally from Algeria: \RL{نيلمار سانتوس دا سيلفا. موهبة جزائرية تصنع أفراح سانتوس البرازيلي}.

\subsection{Viral spread and echo chambers}

The spread of fake news reflects both network dynamics and algorithmic amplification \cite{Cinelli2021,Trnberg2018EchoCA,Alatawi2021ASO,Cota2019QuantifyingEC,Wang2024InsideTE}. Clusters of like-minded users reinforce existing beliefs through homophily, while platform algorithms tend to prioritize engaging or sensational content. This dual reinforcement mechanism accelerates content diffusion and contributes to the formation of localized echo chambers, making misinformation more persistent and difficult to counter.

In this work, we operationalize and quantify echo-chamber intensity at the level of individual fake-news narratives.

We restrict the analysis to entries labeled as \textit{Fake} and their associated \textit{FakeComment} entries sharing the same subject. We note that the \textit{subject} field is used as a proxy for narrative clustering. For each subject $s$, let $E_s=\{e_1,\dots,e_n\}$ denote the corresponding set of entries, where each entry contains message text, user identifier, and publication date.

\paragraph{Semantic similarity}
To quantify textual redundancy within each narrative, messages are encoded using the multilingual sentence-transformer model \texttt{paraphrase-multi\-lingual-MiniLM-L12-v2} \citep{reimers-2019-sentence-bert}. Let $x_i \in \mathbb{R}^d$ denote the embedding of message $e_i$. Here, $d$ denotes the dimensionality of the embedding space. Pairwise semantic similarity is computed using cosine similarity:
\begin{equation}
\mathrm{sim}(e_i,e_j)=\frac{x_i^\top x_j}{\|x_i\|\,\|x_j\|}.
\end{equation}
The semantic similarity score for subject $s$ is defined as the average pairwise similarity:
\begin{equation}
S_s=\frac{2}{n(n-1)}\sum_{1\le i<j\le n}\mathrm{sim}(e_i,e_j).
\end{equation}
Higher values of $S_s$ indicate stronger linguistic convergence, where users tend to reproduce similar formulations or narrative frames.

\paragraph{Author concentration}
To measure whether dissemination is dominated by a small number of users, we compute the Herfindahl--Hirschman Index (HHI) \citep{hirschman1964national, rhoades1993herfindahl}. Let $c_u$ denote the number of entries produced by user $u$ in subject $s$, and $N=\sum_u c_u$ the total number of entries. The concentration index is defined as:
\begin{equation}
H_s=\sum_{u=1}^{m}\left(\frac{c_u}{N}\right)^2,
\end{equation}
where $m$ is the number of distinct users. Values of $H_s$ close to 1 indicate highly centralized diffusion, while lower values correspond to more distributed participation across users.

\paragraph{Temporal burstiness}
To capture the speed of diffusion, we measure the proportion of entries published within 24 hours of the first observed message in a subject. Let $d_{\min}$ be the earliest publication date in $E_s$. We define, the burstiness score is defined as:
\begin{equation}
B_s=\frac{\#\{e_i \in E_s : d_i \le d_{\min}+24\text{h}\}}{n}.
\end{equation}
where $d_i$ is the timestamp of the message. Higher values of $B_s$ indicate rapid and concentrated diffusion, characteristic of viral or coordinated spreading patterns.

\paragraph{Echo Chamber Index}
We define the Echo Chamber Index (ECI) as a weighted combination of the three indicators:
\begin{equation}
\mathrm{ECI}_s = \alpha\,S_s + \beta\,H_s + \gamma\,B_s,
\end{equation}
with $\alpha=0.5$, $\beta=0.3$, and $\gamma=0.2$.

This formulation prioritizes semantic redundancy while also accounting for author concentration and temporal dynamics.

\paragraph{Qualitative interpretation}
We map the ECI into qualitative categories as follows:
\begin{equation}
\mathrm{EchoLevel}(s)=
\begin{cases}
\text{Weak}, & \text{if } \mathrm{ECI}_s < 0.30, \\
\text{Moderate}, & \text{if } 0.30 \le \mathrm{ECI}_s < 0.60, \\
\text{Strong}, & \text{if } \mathrm{ECI}_s \ge 0.60.
\end{cases}
\end{equation}

This classification allows us to distinguish between weak, moderate, and strong echo-chamber dynamics at the narrative level.

\paragraph{Results}
\label{sec:results}

\begin{table}[htbp]
\centering
\small
\begin{tabular}{lrrrrrl}
\toprule
\textbf{Subject} & \textbf{Entries} & \textbf{Users} & \textbf{Semantic} & \textbf{HHI} & \textbf{Burst} & \textbf{ECI} \\
\midrule
D\'EC\`ES TOUFIK (Death of Toufik) & 2 & 1 & 0.73 & 1.00 & 1.00 & 0.86 \\

TUNISIENS EXPULSÉS GUINÉE & 2 & 1 & 0.60 & 1.00 & 1.00 & 0.80 \\
\,\,\,(Tunisians Expelled from Guinea) &  &  & & & & \\
FRANCE ALGÉRIE (France Algeria) & 2 & 2 & 0.87 & 0.50 & 1.00 & 0.78 \\
DECES DE TOUS LES MALADES  & 2 & 2 & 0.86 & 0.50 & 1.00 & 0.78 \\
\,\,\,HOPITAL (Death of All  &  & &  &  & & \\
\,\,\,Hospital Patients) &  & &  &  & & \\
BEBES PALESTINIENS ARMEE&  3 & 3 & 0.94 & 0.33 & 1.00 & 0.77 \\
\,\,\,ISRAELIENNE (Palestinian Babies &  &  &  &  &  \\
\,\,\,and the Israeli Army) &  &  &  &  &  \\
D\'EC\`ES BOUTEFLIKA (Death of  & 301 & 5 & 0.51 & 0.97 & 0.006 & 0.55 \\
\,\,\,Bouteflika) &  & &  &  & & \\
\bottomrule
\end{tabular}
\caption{Selected fake-news narratives and their echo-chamber scores.}
\label{tab:echo_examples}
\end{table}

The results in Table \ref{tab:echo_examples} indicate that strong echo-chamber structures are typically characterized by a combination of high semantic similarity, high author concentration, and rapid temporal bursts. For instance, \textit{D\'EC\`ES TOUFIK} (death of Toufik) contains only 2 entries produced by a single user, with a semantic similarity of 0.7322, an HHI of 1, and a burstiness of 1. Its resulting ECI is 0.8661, placing it in the \textit{Forte} category. Similarly, \textit{TUNISIENS EXPULSÉS GUINÉE} (Tunisians expelled from Guinea) and \textit{FRANCE ALGÉRIE} (France Algeria) also exhibit high ECI values, reflecting highly concentrated and rapidly repeated narratives.

The subject \textit{BEBES PALESTINIENS ARMEE ISRAELIENNE} (Palestinian babies, Israeli army) also shows strong echo-chamber behavior, with 3 entries from 3 distinct users and a very high semantic similarity (0.947). Despite a more distributed authorship structure, the ECI remains high (0.7735), indicating that semantic repetition and temporal concentration can compensate for lower author centralization.

In contrast, subjects such as \textit{TEBBOUNE REMERCIE MAROC ENVOIE AVIONS INCENDIE} (Tebboune thanks Morocco for sending canadairs), \textit{D\'EC\`ES BOUTEFLIKA} (death of Bouteflika), and \textit{PRESIDENT TUNISIEN ABSTENTION CESSEZ FEU} (The Tunisian president abstains from voting on the ceasefire) fall into the Moderate category. For example, \textit{D\'EC\`ES BOUTEFLIKA} (death of Bouteflika) contains 301 entries and 5 users, but exhibits moderate semantic similarity (0.5192) and very low burstiness (0.0066), resulting in an ECI of 0.553. This suggests that high message volume alone does not imply strong echo-chamber behavior when diffusion is temporally dispersed and semantically diverse.

Overall, the ECI successfully differentiates between tightly clustered misinformation loops and more diffuse narrative structures. Small, highly repetitive, and rapidly spreading narratives tend to exhibit the strongest echo-chamber effects, while larger narratives may remain only moderately structured if they lack temporal synchronization or semantic redundancy. These results support the usefulness of combining semantic similarity, author concentration, and temporal burstiness as complementary dimensions of misinformation diffusion.

\section{Correlation between fake news and social media metrics}
In the following we will use some classical social media metrics to analyse BOUTEF.
\subsection{Followers and engagement}
The relationship between fake news and social media metrics reveals significant patterns in both statistical correlations and engagement disparities. Quantitative analysis indicates a positive relationship between the number of followers a fake news creator has and the engagement rate of their posts. 

In this study, the engagement rate for a user is calculated as follows:
\[
\text{Engagement Rate (\%)} = \frac{\text{Total Comments}}{\text{Total Followers}} \times 100
\]
This measure expresses the proportion of followers who interacted with a post through comments.

Table \ref{tab:fake-engagement} shows a highly unequal engagement landscape across fake-news accounts. The top two accounts (Users 1 and 2) reach exceptionally high engagement rates (96.00\% and 95.60\%), indicating near one-to-one interaction between follower base and comment volume for the analyzed posts.
\begin{table}[h!]
\centering
\begin{tabular}{|l|r|r|r|r|}
\hline
\textbf{User} & \textbf{Fake Posts} & \textbf{Total Followers} & \textbf{Total Comments} & \textbf{Engagement Rate (\%)} \\
\hline
User 1 & 1 & 10000 & 9600 & 96.00 \\
\hline
User 2 & 1 & 22000 & 21033 & 95.60 \\
\hline
User 3 & 1 & 686 & 307 & 44.75 \\
\hline
User 4 & 1 & 686 & 305 & 44.46 \\
\hline
User 5 & 3 & 17100 & 5700 & 33.33 \\
\hline
User 6 & 3 & 924 & 306 & 33.12 \\
\hline
User 7 & 1 & 16400 & 4875 & 29.73 \\
\hline
User 8 & 1 & 4012 & 1046 & 26.07 \\
\hline
User 9 & 1 & 3500 & 530 & 15.14 \\
\hline
User 10&	1&	3840&	504&	13,13\\
\hline
\end{tabular}
\caption{Top 9 Fake News Accounts Ranked by Engagement.}
\label{tab:fake-engagement}
\end{table}
 A second tier (approximately 26\%--45\%) includes Users 3, 4, 5, 6, 7, and 8, suggesting substantial but less extreme audience activation. The lowest rates are observed for Users 9 and 10 (15.14\% and about 13.13\%), which still represent non-negligible interaction levels. Overall, these results suggest that virality is concentrated in a small number of high-performing accounts, while most accounts generate moderate engagement; this concentration can amplify the visibility of fake narratives when top accounts publish even a small number of posts.

Political fake news exhibits the strongest correlations between follower counts and engagement levels, suggesting that creators with larger audiences can amplify political misinformation more effectively. This highlights the potential for widespread influence and manipulation via social media platforms.

A concrete example occurred on February 25, 2025, when a wave of misinformation spread across social media, centered on a fabricated report concerning Algerian President Abdelmadjid Tebboune:
\RL{عاجل وفاة الرئيس الجزائري عبد المجيد تبون في ظروف غامضة} \\
(Urgent: The death of Algerian President Abdelmadjid Tebboune under mysterious circumstances).
The false narrative claimed his sudden and mysterious death, proliferating rapidly, particularly on TikTok. Two influencers, with followings of 50,700 and 46,500, played a key role in disseminating this misleading content. The falsity was quickly exposed when President Tebboune appeared in subsequent media engagements.

The influencers’ actions demonstrated two main motivations: undermining the Algerian government by suggesting a cover-up, and exploiting the viral nature of sensational news for personal gain. This incident illustrates how easily fake news spreads. The rapid dissemination was fueled by TikTok's algorithm, which prioritizes viral content, and by the use of sensationalist headlines designed to capture attention. Public anxiety and fear also contributed to unverified sharing. The case underscores the power of influencers in spreading disinformation and the potential for political manipulation, as well as the resulting erosion of public trust.

\subsection{Influence of social media algorithms}
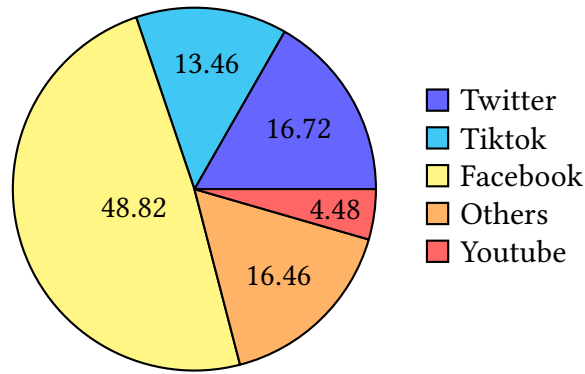
\begin{figure}[H]
\begin{tikzpicture}[scale=0.8]
\pie[text=legend, sum=auto, after number=, radius=3]{
16.72/Twitter,
13.46/Tiktok,
48.82/Facebook,
16.46/Others,
4.48/Youtube
}
\end{tikzpicture}
\caption{Distribution of narratives in the BOUTEF-FAKE part.}
\label{platforms}
\end{figure}

Platform-specific trends also shape the spread of fake news. TikTok and YouTube rely heavily on video content, making them susceptible to video-based misinformation, such as manipulated footage. In contrast, Facebook shows a higher prevalence of text-based misinformation, including fabricated articles and misleading posts. Figure \ref{platforms} illustrates that the majority of the fake news within BOUTEF data was scraped from Facebook, with most posts accompanied by images or videos to enhance engagement.
These patterns highlight the need for tailored strategies that consider each platform's unique content formats and dissemination dynamics.

\begin{table}[H]
\centering
\begin{tabular}{|l|r|}
\hline
\textbf{Platform} & \textbf{Proportion (\%)} \\
\hline
Youtube  & 89.07 \\
Facebook  & 5.63 \\
Tiktok  & 2.51 \\
Twitter  & 1.76 \\
Others  & 1.02 \\
\hline
\end{tabular}
\caption{Distribution of Narratives Across Social Media Platforms in the whole BOUTEF Dataset}
\label{tab:platforms}
\end{table}
Table \ref{tab:platforms} summarizes the platform distribution for the full BOUTEF corpus, aggregating all narrative categories (e.g., Fake, FakeComment, NoFake, Genuine, and related content types). The distribution is highly imbalanced: YouTube alone accounts for 89.07\% of the dataset, while Facebook (5.63\%), TikTok (2.51\%), Twitter (1.76\%), and Others (1.02\%) represent much smaller shares. This indicates that analyses performed on the complete corpus are strongly influenced by YouTube-driven content dynamics; therefore, platform-aware normalization or stratified evaluation is necessary to avoid biased conclusions when comparing narrative behaviors across sources. This imbalance is largely explained by the data collection process: automatic scraping was performed predominantly on YouTube.

\section{Geographical comparison analysis: Algeria vs. Tunisia}

\begin{figure}[H]
    \centering
    \includegraphics[width=\textwidth]{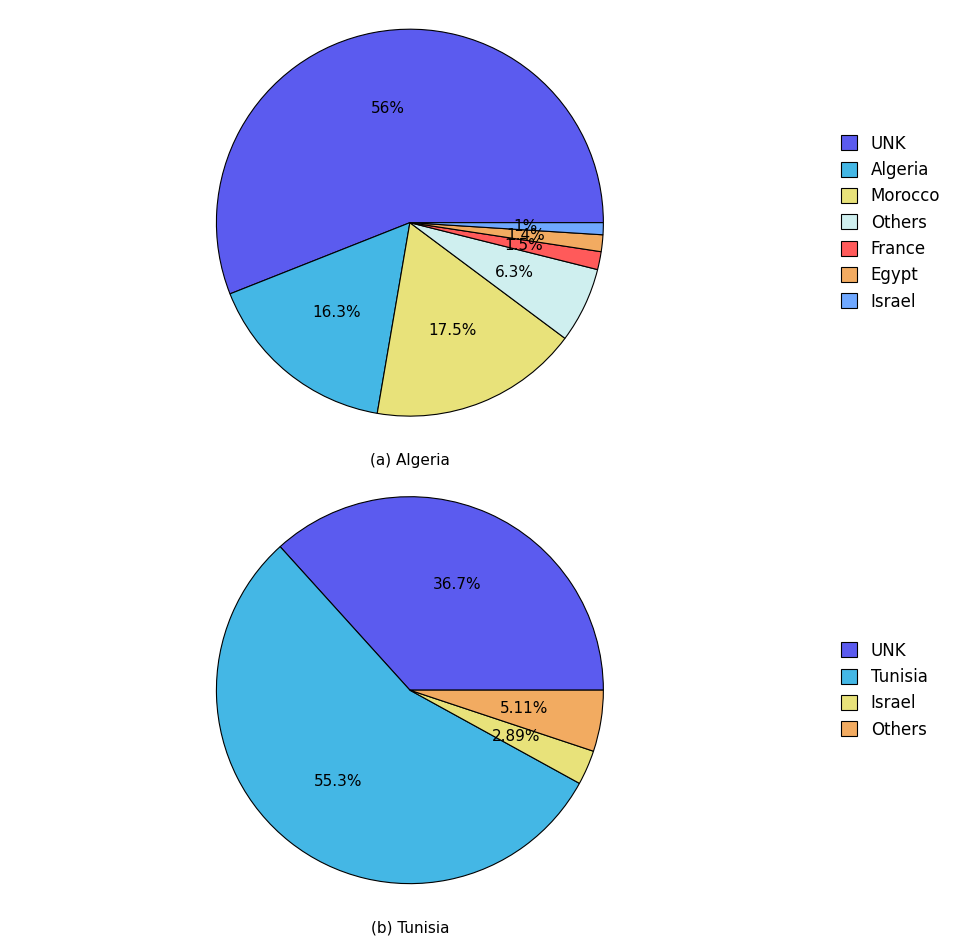}
    \caption{Regional distribution in fake news for Algeria and Tunisia.}
    \label{CountCombined}
\end{figure}

The origin of fake news writer in the BOUTEF corpus differs markedly if the topic concerns Algeria or Tunisia, as shown in Figure \ref{CountCombined}. In both cases, a substantial share of the material remains unattributed (UNK), which already indicates the difficulty of tracing misinformation sources with precision. However, beyond this common uncertainty, the two countries display distinct origin profiles.

For Algeria, the distribution is more externally oriented and geographically diversified. UNK represents the largest share (56\%), but among the identified origins, Morocco accounts for 17.5\% and Algeria itself for 16.3\%. Smaller proportions come from France (1.5\%), Egypt (1.4\%), Israel (1\%), and other origins (6.3\%). This pattern suggests that fake news targeting Algeria is not driven primarily by domestic sources alone; rather, it reflects a mixture of internal circulation and cross-border regional influence, especially from the Maghreb.

Tunisia shows a different configuration. While UNK still represents a substantial proportion (36.7\%), the dominant identified origin is Tunisia itself (55.3\%), whereas Israel accounts for 2.89\% and the remaining 5.11\% falls into other origins. Compared with Algeria, Tunisian fake news therefore appears much more internally generated or domestically relayed. This stronger domestic concentration may reflect the centrality of national political and social issues in Tunisia's misinformation ecosystem.

Overall, the comparison suggests that fake news affecting Algeria has a more heterogeneous and transnational origin structure, whereas fake news affecting Tunisia is more strongly rooted in internal dissemination. At the same time, the high proportion of UNK in both countries remains an important limitation and calls for caution when interpreting source attribution. These findings nonetheless highlight the value of country-specific analyses for understanding how misinformation circulates across different national contexts.

\section{Measuring lexical richness and language diversity across message types}

To quantify the linguistic diversity of each message type in the BOUTEF corpus,
we computed four complementary metrics over the five content categories:
\textit{Fake}, \textit{FakeComment}, \textit{NoFake}, \textit{Genuine},
and \textit{GenuineComment}.
These measures operate at two levels: the \textit{lexical} level, capturing
the richness and distribution of the vocabulary within each group, and the
\textit{language} level, capturing the diversity of the declared languages
across entries.

\subsection{Normalised Shannon entropy}

The Normalised Shannon entropy $H_{\text{norm}}$ measures how
evenly the vocabulary is distributed across tokens.
Let $f_i$ denote the frequency of word $i$ in a group of $N$ tokens and
$V$ the vocabulary.
The raw Shannon entropy is
\begin{equation}
  H = -\sum_{i=1}^{V} p_i \log_2 p_i,
  \qquad p_i = \frac{f_i}{N},
  \label{eq:shannon}
\end{equation}
and its normalised form, which enables comparisons across groups of different
sizes, is
\begin{equation}
  H_{\text{norm}} = \frac{H}{\log_2 V} \in [0,1].
  \label{eq:hnorm}
\end{equation}
A value close to 1 indicates a highly balanced vocabulary
(each word used with similar frequency), while a value close to 0 indicates
that a small number of words dominate.
\subsection{Yule's K measure}
Yule's $K$ \cite{YuleK} provides a size-robust measure of lexical
richness that is largely independent of text length.
Let $V_m$ denote the number of word appearing exactly $m$ times.
Then
\begin{equation}
  K = 10^4 \times \frac{\displaystyle\sum_{m=1}^{\infty} V_m \cdot m^2 - N}{N^2}.
  \label{eq:yulek}
\end{equation}
where $N$ is the number of tokens in the message as for Shannon entropy. A \textit{higher} $K$ value indicates a \textit{poorer} vocabulary
(fewer distinct types relative to repetitions), while a lower $K$ indicates
greater lexical richness.
\subsection{Language entropy}
The language entropy $E_{\text{lang}}$ applies Shannon's formula
(Eq.~\ref{eq:shannon}) to the distribution of the declared \texttt{LANGUAGE}
field rather than to the word distribution:
where $p_i$ is the proportion of entries in language $i$ within a group.
A high $E_{\text{lang}}$ reflects a linguistically heterogeneous group,
while a low value indicates near-monolingual production.

\subsection{Results}

Table~\ref{tab:lexical_stats} presents the corpus-level statistics for each
content type. Several observations emerge from these figures.

\begin{table}[!h]
\centering
\begin{tabular}{lrrrrrrr}
\toprule
Type & $N_{\text{Post}}$ & $N_{\text{Tok}}$ & Voc
     & $Yule K$ & $H_{\text{norm}}$ & $E_{\text{lang}}$ \\
\midrule
Fake           &    891 &    20{,}662 &  8{,}283 & 18.07 & 0.90 & 1.53 \\
FakeComment    &  5{,}142 &    67{,}803 & 23{,}241 & 14.40 & 0.86 & 2.56 \\
NoFake         &    448 &    30{,}021 &  9{,}723 & 20.01 & 0.87 & 1.30 \\
Genuine        &    309 &     6{,}803 &  3{,}332 & 26.07 & 0.90 & 2.02 \\
GenuineComment & 50{,}321 & 908{,}627 & 90{,}439 & 15.91 & 0.76 & 1.75 \\
\bottomrule
\end{tabular}
\caption{Summary of lexical diversity statistics by message type.
  $N_{\text{Post}}$: number of messages;
  $N_{\text{Tok}}$: total tokens;
  Voc: vocabulary size (unique types);
  $YuleK$: Yule's $K$;
  $H_{\text{norm}}$: normalised Shannon entropy;
  $E_{\text{lang}}$: language entropy.}
\label{tab:lexical_stats}
\end{table}

\textbf{Lexical richness.}
Yule's $K$ reveals a clear ordering of vocabulary richness.
\textit{Genuine} content scores the highest $K$ value ($K = 26.07$),
indicating the \textit{most repetitive} vocabulary.
This is consistent with the nature of institutional and journalistic output,
which relies on a recurrent domain-specific lexicon
(diplomatic, economic, and political terminology).
At the other extreme, \textit{FakeComment} achieves the lowest $K$ ($K = 14.40$),
reflecting the aggregated lexical contributions of thousands of users each
bringing their own dialectal expressions, colloquialisms, and emotionally
charged vocabulary.
\textit{NoFake} debunking content ($K = 20.01$) falls between these poles,
combining a standardised fact-checking lexicon with some topical variability.

\textbf{Normalised lexical entropy.}
Four of the five groups cluster in a narrow range of $H_{\text{norm}} \in [0.86,
0.90]$, indicating that their vocabularies are distributed comparably across
types. Notably, \textit{Fake} and \textit{Genuine} share the same value
($H_{\text{norm}} = 0.90$), while \textit{NoFake} ($0.87$) and
\textit{FakeComment} ($0.86$) are only marginally lower.
This convergence between \textit{Fake} and verified content types is
particularly significant: it suggests that fake news creators deliberately adopt
a lexical distribution that mimics legitimate reporting, thereby increasing
apparent credibility. This observation supports the findings discussed in
Section~5 on media imitation as a manipulation strategy.
\textit{GenuineComment} stands apart with the lowest $H_{\text{norm}} = 0.76$,
a consequence of Zipf's law at scale: with over 900{,}000 tokens,
a handful of high-frequency items (religious expressions, national identifiers)
compress the entropy substantially.

\textbf{Language entropy.}
$E_{\text{lang}}$ reveals the sharpest contrasts across groups.
\textit{FakeComment} records the highest language entropy ($E_{\text{lang}} = 2.56$),
confirming that public reactions to disinformation in North Africa are
fundamentally multilingual, mixing MSA, Algerian and Tunisian dialects,
Arabizi, French, and English within the same comment thread.
Conversely, \textit{NoFake} debunking content has the lowest language
entropy ($E_{\text{lang}} = 1.30$), reflecting the deliberate choice of
fact-checking organisations to publish predominantly in MSA and French in
order to maximise cross-national reach and institutional legitimacy.
\textit{Fake} content itself is more linguistically homogeneous
($E_{\text{lang}} = 1.53$) than the comments it generates, suggesting that
disinformation producers strategically select their language, whereas audience
reactions are spontaneous and linguistically diverse.

\paragraph{Summary.}
Taken together, these results show that lexical diversity alone is not a
reliable signal for distinguishing genuine from fabricated content.
Rather, it is the \textit{pattern} of metrics high language entropy in
comments, convergent $H_{\text{norm}}$ between Fake and Genuine, elevated
$K$ in institutional output that characterises the different communicative
registers of the corpus.
These findings motivate the inclusion of diversity-based features, particularly
$H_{\text{norm}}$ and $E_{\text{lang}}$, alongside traditional lexical features
in future misinformation detection models trained on BOUTEF.

\begin{figure}[!h]
\centering
\includegraphics[scale=0.5]{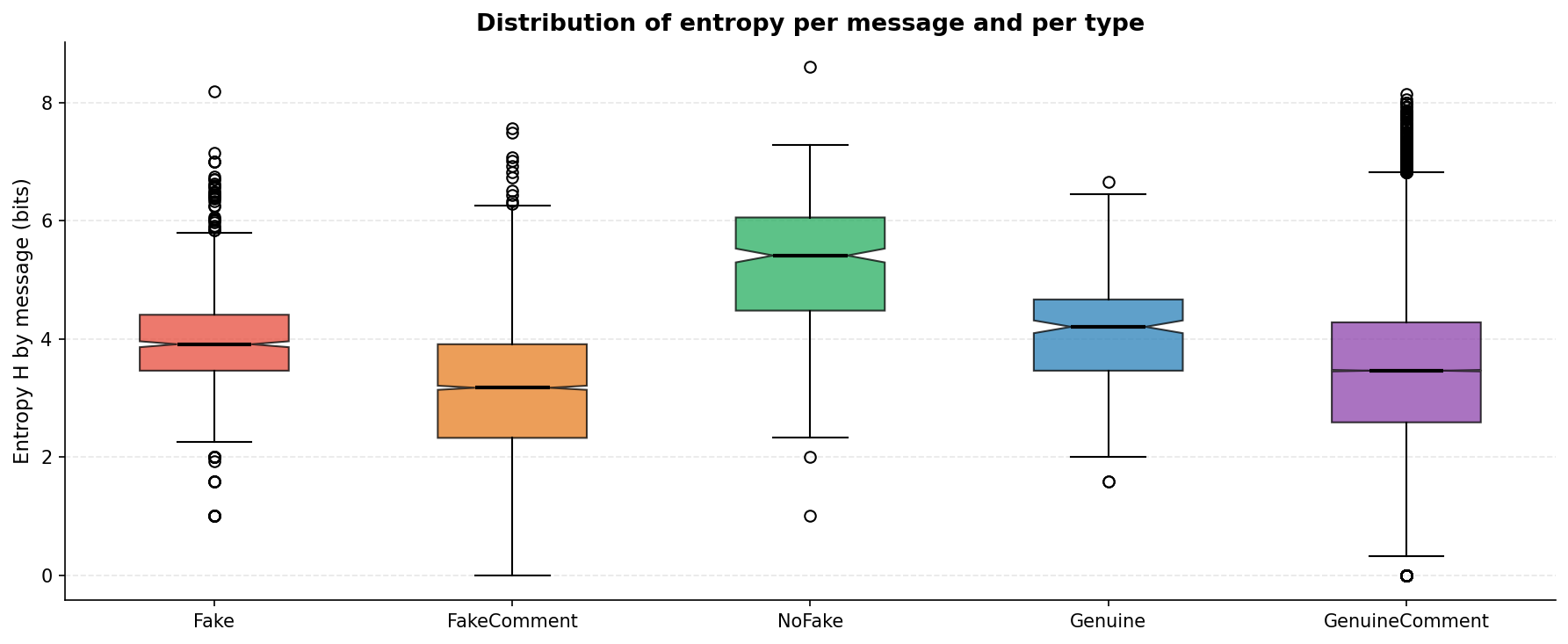}
\caption{Distribution of entropy per message and per type.}
\label{fig:boxplot_entropy}
\end{figure}

Figure~\ref{fig:boxplot_entropy} displays the per-message Shannon entropy
distributions across the five content types in BOUTEF.

The most striking observation is that \textit{NoFake} debunking content
stands entirely apart from all other groups, combining the highest median
entropy ($\tilde{H} \approx 5.3$ bits) with the narrowest interquartile
range ($\text{IQR} \approx 1.2$ bits), reflecting both the lexical richness
and the stylistic consistency of institutional fact-checking discourse.
By contrast, \textit{FakeComment} and \textit{GenuineComment} share
an almost identical distribution profile, low median ($\approx 3.2$--$3.4$
bits), whiskers reaching zero, and a heavy upper tail of outliers, confirming
that it is the \textit{format} (short reactive comment) rather than the
\textit{veracity} of the content that governs lexical diversity in user
reactions.
Remarkably, \textit{Fake} and \textit{Genuine} messages are
virtually indistinguishable, with overlapping boxes and near-identical
medians ($\approx 3.9$ and $4.1$ bits respectively), a result consistent
with the convergent $H_{\text{norm}}$ values reported in
Table~\ref{tab:lexical_stats}.
This overlap indicates that fake news creators successfully mimic the
lexical distribution of legitimate reporting, rendering per-message entropy
alone an insufficient discriminating feature for automated detection.
The notch analysis corroborates these findings: the confidence intervals
around the \textit{NoFake} median do not overlap with any other group,
whereas those of \textit{Fake} and \textit{Genuine} partially coincide,
suggesting that the former is the only type statistically separable
from the rest at the 5\% level.
Taken together, these results argue for moving beyond purely lexical
features toward multimodal and contextual representations in future
misinformation detection models trained on BOUTEF.

\section{Conclusion}
The analysis of the BOUTEF corpus highlights the complex and dynamic nature of fake news. Misinformation is not static: its thematic and temporal evolution mirrors ongoing events, with surges observed during crises, particularly in political and health-related domains. This adaptability shows the need for continuous monitoring and rapid responses.  

The strategies employed by misinformation creators reveal a high degree of sophistication. From sensationalist headlines to manipulated imagery, these rhetorical and visual techniques are deliberately designed to bypass critical thinking and exploit cognitive biases. These findings highlight the importance of media literacy programs that build critical thinking skills.

Linguistic and demographic variability further shape the dissemination of fake news. The widespread use of Arabizi, dialects, and code-switching demonstrates how misinformation leverages generational and cultural communication practices to enhance virality across diverse audiences. These observations point to the need for culturally sensitive and linguistically inclusive approaches to counter misinformation.

The role of social media platforms is central in amplifying fake news. Algorithm-driven promotion and influencer activity create powerful vectors for rapid diffusion, while echo chambers limit corrective debate. Cross-platform collaboration and algorithmic accountability are therefore essential components of mitigation strategies.  

Finally, the comparative analysis of Algeria and Tunisia shows both shared challenges and national specificities. While both face a prevalence of political misinformation, Algeria’s narratives often draw on historical and identity-based references, whereas Tunisia’s focus on democratic transition and reform reflects its evolving sociopolitical context. Tailored strategies are thus required to address the distinct vulnerabilities of each national information ecosystem.  

This study nonetheless has limitations, including the temporal scope of the corpus, potential representativeness issues, and challenges in source attribution (e.g., the UNK category). Consequently, our findings should not be generalized without caution, especially because data collection did not always provide access to all contextual or account-level information needed for exhaustive interpretation. Future research could extend the analysis to other Maghreb countries, integrate multimodal data (e.g.  sounds, videos), and evaluate the effectiveness of counter-narratives.  

Overall, the findings contribute to a deeper understanding of the multifaceted dynamics of fake news and provide empirical evidence to inform media literacy initiatives, fact-checking efforts, and policy frameworks aimed at building more resilient and informed societies. The BOUTEF corpus is publicly available at: \url{https://huggingface.co/datasets/TeamSmart/BOUTEF}.

\section*{ACKNOWLEDGEMENTS}
We acknowledge the AID and ANR for their invaluable financial support, which made this research for TRADEF project endeavor possible. The guidance and insights provided were instrumental in the successful execution of our study.

\bibliographystyle{ACM-Reference-Format}
\bibliography{ref}
\newpage
\onecolumn
\appendix

\section*{Appendix A: Fake News Topics in BOUTEF}
\label{A}

\begin{itemize}
\item \textbf{ADMINISTRATION}: Issues related to public administration, services,etc.
\item \textbf{ARMY}: Topics involving the military or armed forces.
\item \textbf{CATASTROPHE}: Stories about natural or man-made disasters.
\item \textbf{CELEBRITY DEATH}: Reports about the passing of famous individuals.
\item \textbf{CELEBRITY ILLNESS}: Reports about illnesses affecting celebrities.
\item \textbf{CONSPIRACY}: Claims involving secret plots or conspiracies.
\item \textbf{CORONA}: Information about the coronavirus (COVID-19).
\item \textbf{CULTURE}: Issues tied to cultural practices or traditions.
\item \textbf{DEBT}: Claims concerning public, private, or international debt.
\item \textbf{DEMOGRAPHY}: Topics related to population structure.
\item \textbf{DENIGRATION}: Content aimed at discrediting individuals or groups.
\item \textbf{DEVOTION}: Acts of sacrifice or religious devotion.
\item \textbf{DIPLOMACY}: Stories involving diplomatic relations, negotiations, or representation.
\item \textbf{ECONOMY}: Matters concerning financial systems or economic policies.
\item \textbf{EDUCATION}: Issues concerning schools, universities, teaching, learning, or education policy.
\item \textbf{ELECTION}: Claims about electoral processes or outcomes.
\item \textbf{ENERGY}: Topics related to electricity, fuel, natural resources, or energy policy.
\item \textbf{ENVIRONEMENT}: Issues concerning the environment, pollution, climate.
\item \textbf{ESPIONAGE}: Stories about spying or intelligence activities.
\item \textbf{FEAR}: Content designed to incite panic or anxiety.
\item \textbf{FOOD}: Stories related to food supply, consumption, safety, or agricultural products.
\item \textbf{FOREIGN POLICY}: Claims about a state's external strategy, alliances.
\item \textbf{HEALTH}: Stories about medical or public health issues.
\item \textbf{HERITAGE}: Content about cultural or national heritage.
\item \textbf{HIRAK}: Content tied to the Hirak protest movement.
\item \textbf{HISTORY}: Stories about historical events or figures.
\item \textbf{HUMAN RIGHTS}: Issues involving civil liberties, equality, dignity, etc.
\item \textbf{HUMILIATION}: Content designed to shame or embarrass.
\item \textbf{IMMIGRATION}: Topics about migration or immigrants.
\item \textbf{INJUSTICE}: Claims about unfair treatment or inequality.
\item \textbf{JUSTICE}: Topics concerning courts, trials, legal decisions, or the judicial system.
\item \textbf{LANGUAGE}: Issues related to linguistic matters.
\item \textbf{MALFUNCTION}: Stories about failures or breakdowns.
\item \textbf{NATIONAL ANTHEM}: Issues involving national symbols or anthems.
\item \textbf{PAST}: Narratives tied to historical or nostalgic events.
\item \textbf{PERSONALITY}: Stories about public figures or personalities.
\item \textbf{POLICE}: Reports about law enforcement or policing.
\item \textbf{POLITICS}: Topics involving political systems or figures.
\item \textbf{PRIDE}: Narratives promoting national or cultural pride.
\item \textbf{RACISM}: Claims about racial discrimination or bias.
\item \textbf{REGIONALISM}: Narratives about regional identity or autonomy.
\item \textbf{RELIGION}: Issues tied to religious beliefs or groups.
\item \textbf{SCIENCE}: Claims about scientific research or discoveries.
\item \textbf{SENSATIONALISM}: Exaggerated or sensational stories.
\item \textbf{SEPARATISM}: Narratives about movements for regional independence.
\item \textbf{SPORTS}: Topics involving athletic events or figures.
\item \textbf{SUPERSTITION}: Claims based on myths or superstitions.
\item \textbf{TECHNOLOGY}: Stories about digital tools, innovation, or technical systems.
\item \textbf{TERRORISM}: Stories about acts of terror or extremism.
\item \textbf{THIRD WORLD}: Issues related to developing nations.
\item \textbf{TOURISM}: Topics about travel or the tourism industry.
\item \textbf{TRANSPORTATION}: Topics involving mobility, vehicles or transport services.
\item \textbf{TV INSTITUTION}: Stories about television or media organizations.
\item \textbf{UNTHINKABLE SITUATION}: Reports about shocking or unimaginable events.
\item \textbf{WAR}: Reports about conflicts or wars.
\item \textbf{WORK}: Stories about labor or employment issues.
\end{itemize}
\newpage
\section*{Appendix B: Geographical Distribution of Fake News Origins by Country
}
\label{B}
\begin{table}[htbp]
\centering
\small
\begin{tabular}{l r}
\toprule
\textbf{Country} & \textbf{Frequency} \\
\midrule
Unknown & 461 \\
Tunisia & 139 \\
Morocco & 118 \\
Algeria & 104 \\
Israel & 14 \\
France & 12 \\
Egypt & 10 \\
United Kingdom (GB) & 5 \\
Yemen & 3 \\
Ivory Coast & 3 \\
Palestine & 3 \\
Mali & 2 \\
Syria & 2 \\
Jordan & 2 \\
United Kingdom (UK) & 2 \\
Algerian opposition & 1 \\
Morocco/Algeria & 1 \\
Lebanon & 1 \\
Iraq & 1 \\
Middle East & 1 \\
United States & 1 \\
Gambia & 1 \\
Congo & 1 \\
Opposition & 1 \\
Belgium & 1 \\
Saudi Arabia & 1 \\
\bottomrule
\end{tabular}
\caption{Country distribution in fake news narratives}
\label{tab:country_distribution}
\end{table}

\newpage
\section*{Appendix C: Geographical Distribution of Fake comments Origins by Country
}
\begin{table}[htbp]
\centering
\small
\begin{tabular}{l r}
\toprule
\textbf{Country} & \textbf{Frequency} \\
\midrule
Unknown & 4437 \\
Algeria & 387 \\
Morocco & 205 \\
Tunisia & 43 \\
France & 43 \\
Israel & 5 \\
Palestine & 5 \\
Egypt & 4 \\
Belgium & 3 \\
Syria & 3 \\
United Arab Emirates (UAE) & 1 \\
Nigeria & 1 \\
Ghana & 1 \\
Sudan & 1 \\
Kuwait & 1 \\
United Kingdom (Great Britain) & 1 \\
Yemen & 1 \\
\bottomrule
\end{tabular}
\caption{Country distribution in fakeComments news narratives}
\label{tab:country_distribution}
\end{table}

\appendix

\end{document}